%% file: main.tex
\pdfoutput=1
\documentclass{article} 
\usepackage{iclr2027_conference,times}

\input{math_commands.tex}

\usepackage{hyperref}
\usepackage{url}
\usepackage{marvosym}
\usepackage{graphicx}
\usepackage{wrapfig}
\usepackage[caption=false,font=small]{subfig}
\usepackage{booktabs}
\usepackage{colortbl}
\usepackage{multirow}
\usepackage{tabularx}
\usepackage{fvextra}
\usepackage[breakable]{tcolorbox}

\makeatletter
\renewcommand{\paragraph}{\@startsection{paragraph}{4}{\z@}{0pt}{-1em}{\normalsize\bfseries}}
\makeatother

\title{VTR-Bench: A Systematic Benchmark for Evaluating Visual Text Rendering in Video~Generation}

\author{%
\begin{minipage}[t]{\dimexpr\textwidth-2\tabcolsep\relax}
\raggedright\normalfont
{\normalsize\bfseries
\mbox{Yu Huang\textsuperscript{1}},
\mbox{Jungang Li\textsuperscript{2}},
\mbox{Zhiyuan Wang\textsuperscript{1}},
\mbox{Yonghua Hei\textsuperscript{2}},
\mbox{Song Dai\textsuperscript{2}},
\mbox{Jiayu Yang\textsuperscript{2}},
\mbox{Deyuan Liu\textsuperscript{4}},
\mbox{Xiang Zheng\textsuperscript{3}},
\mbox{Xiaoshuang Shi\textsuperscript{5}},
\mbox{Hao Cheng\textsuperscript{6}},
\mbox{Kaidi Xu\textsuperscript{1,3,\Letter}}\endgraf}
\vspace{0.65em}
{\small
\textsuperscript{1}Department of Data Science, City University of Hong Kong\\
\textsuperscript{2}The Hong Kong University of Science and Technology (Guangzhou)\\
\textsuperscript{3}The Hong Kong Institute of AI for Science, City University of Hong Kong\\
\textsuperscript{4}Westlake University\quad
\textsuperscript{5}University of Electronic Science and Technology of China\\
\textsuperscript{6}The Hong Kong University of Science and Technology\endgraf}
\end{minipage}
}

\iclrfinalcopy 
\begin{document}

\maketitle
\lhead{Preprint}
\begingroup
\renewcommand{\thefootnote}{}
\footnotetext{Email: \href{mailto:yhuang3273-c@my.cityu.edu.hk}{\texttt{yhuang3273-c@my.cityu.edu.hk}}.}
\endgroup
\begingroup
\renewcommand{\thefootnote}{\Letter}
\footnotetext{Corresponds to: \href{mailto:kaidixu@cityu.edu.hk}{\texttt{kaidixu@cityu.edu.hk}}.}
\endgroup

\begin{abstract}

Recent video generation models can produce highly realistic videos from natural language instructions, with visual quality approaching cinematic standards. Existing evaluation benchmarks, however, predominantly assess visual quality, aesthetic appeal and physical plausibility, while paying limited attention to text, an essential medium for conveying information in everyday scenes. A generated video may appear visually compelling and feature lifelike subjects, yet still render the text within the scene incorrectly. To address this overlooked dimension, we introduce \textbf{VTR-Bench}, a systematic benchmark for evaluating the \textbf{V}isual \textbf{T}ext \textbf{R}endering capabilities of video generation models. VTR-Bench situates text within concrete application scenarios, such as advertisements and scientific videos, with 300 carefully constructed prompts spanning five scenario categories. We develop an automated evaluation pipeline with human alignments that separately assesses text fidelity through carrier-specific transcription and scene and motion requirements through a prompt-specific chain of query. Beyond evaluation, we introduce a \textbf{Keyframe-Guided Agentic Framework} in which a Director agent coordinates image and video generation with visual evaluation, guiding iterative refinement and candidate selection through visual feedback. Experiments on 11 state-of-the-art models reveal widespread difficulties in accurately rendering scene text, with the best-performing model recording an overall word error rate (WER) of 0.250. We further analyze text rendering failures to characterize the challenges faced by current video generation models. These findings highlight visual text rendering as a key challenge for video generation and demonstrate a practical path toward improvement. Code is available at \url{https://github.com/hardenyu21/VTR-Bench}.

\end{abstract}

\section{Introduction}
\input{Sections/introduction}

\section{Related Work}
\input{Sections/related_work}

\section{VTR-Bench}
\input{Sections/VTR-Bench}

\section{Experiment}
\input{Sections/experiments}

\input{Sections/conclusion}

\bibliography{iclr2027_conference}
\bibliographystyle{iclr2027_conference}

\appendix
\section*{Appendix}
\input{Sections/Appendix/more_related_work}

\input{Sections/Appendix/benchmark}
\input{Sections/Appendix/evaluator_results}

\end{document}

%% file: math_commands.tex
\usepackage{amsmath,amsfonts,bm}

\def\eqref#1{equation~\ref{#1}}

\def\1{\bm{1}}

\DeclareMathAlphabet{\mathsfit}{\encodingdefault}{\sfdefault}{m}{sl}
\SetMathAlphabet{\mathsfit}{bold}{\encodingdefault}{\sfdefault}{bx}{n}



%% file: Sections/introduction.tex
\vspace{-4mm}
\AddToHookNext{cmd/@cflt/before}{\edef\introSavedTextFloatSep{\the\textfloatsep}\setlength{\textfloatsep}{20pt}}
\AddToHookNext{cmd/@cflt/after}{\setlength{\textfloatsep}{\introSavedTextFloatSep}}
\begin{figure}[h]
    \setlength{\abovecaptionskip}{4pt}
    \centering
    \includegraphics[width=\linewidth]{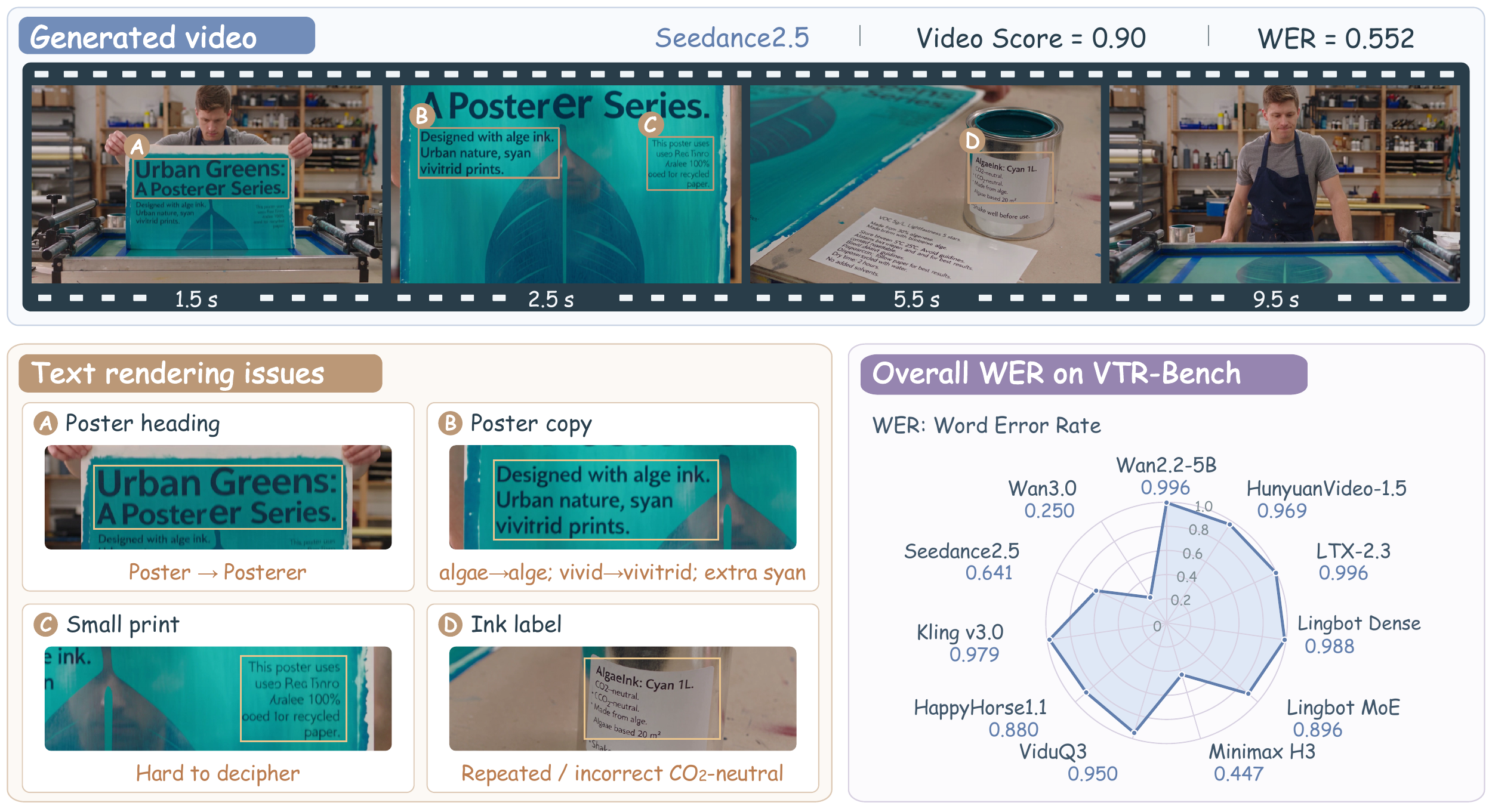}
    \caption{Visual text rendering remains challenging for current video generation models, even in visually convincing videos. Top: a video generated by Seedance2.5. Video Score measures the proportion of satisfied requirements in a 20-question checklist, and WER measures word error rate against the reference text. Bottom left: four text rendering issues in the video. Bottom right: overall WER of 11 video generation models on VTR-Bench.}
    \label{fig:intro}
    \vspace{-4.5mm}
\end{figure}

Recent advances in video generation models have enabled the synthesis of highly realistic videos, with visual quality approaching cinematic standards~\citep{kuaishou2024kling,vidu2026q3,happyhorse2026,minimax2026h3,seedance202625,wanai2026wan3}. However, a convincing visual appearance does not necessarily mean that the text within a scene is correct~\citep{liu2024textanimator,liu2025videotextpreservation}. This distinction matters in applications such as advertisements, scientific demonstrations, and user interfaces, where text conveys essential information through product introduction, numerical values, and instructions~\citep{guo2025t2vtextbench}. In these settings, incorrectly rendered words or symbols can change the intended message, making textual accuracy essential to the usefulness of the generated video. These practical demands motivate a crucial question: \textbf{\textit{can video generation models render the required text accurately?}} As illustrated in Figure~\ref{fig:intro}, an approximately 10-second video contains misspelled words, repeated and incorrect text, and largely illegible passages. Although it achieves a Video Score of 0.90 on a checklist of scene and motion requirements, its word error rate (WER)~\citep{klakow2002testing} reaches 0.552. This example highlights a persistent challenge for current video generation models: accurately rendering visual text even when the surrounding scene and motion requirements are well satisfied.

Existing video generation benchmarks primarily assess visual quality, prompt alignment, compositionality, and physical plausibility~\citep{huang2024vbench,meng2024towards,sun2025t2v,han2025video,zheng2025vbench2,bansal2025videophy,bansal2026videophy}. Benchmarks that explicitly assess rendered text, including EvalCrafter~\citep{liu2024evalcrafter}, T2VTextBench~\citep{guo2025t2vtextbench}, and AVGen-Bench~\citep{zhou2026avgen}, primarily cover short textual targets, with limited coverage of longer passages. In addition, T2VTextBench relies entirely on human evaluation, making repeated model evaluation labor-intensive. Scalable evaluation of longer scene text remains underexplored.

To address this gap, we introduce \textbf{VTR-Bench}, a systematic benchmark for evaluating \textbf{V}isual \textbf{T}ext \textbf{R}endering in video generation. Its 300 carefully constructed prompts span advertising, science, user interfaces, culture, and daily life. Each scene contains multiple textual targets, from short labels to extended passages, with reference strings and carrier annotations specifying what should appear and where. To ground these textual requirements in coherent video scenarios, we adopt a multi-stage construction pipeline with human review. To enable scalable assessment, we develop an automated pipeline that decouples text fidelity from video requirements. A vision--language model transcribes text from specified carriers for comparison with reference strings using WER. In parallel, a prompt-specific chain of query containing 20 questions assesses scene and motion requirements to produce Video Score. Both evaluation branches are validated with human alignments.

Evaluation of 11 state-of-the-art video generation models reveals substantial text errors across both open-source and proprietary models, with the lowest overall WER at 0.250. Moreover, models with similar Video Scores exhibit markedly different text fidelity, separating adherence to scene and motion requirements from the correctness of rendered text. To improve visual text rendering, we develop a \textbf{Keyframe-Guided Agentic Framework} whose Director agent coordinates image and video generation with visual evaluation, guiding iterative refinement and candidate selection through visual feedback. Compared with direct generation using Minimax H3~\citep{minimax2026h3}, our framework reduces overall WER by 32.5\% and improves Video Score. The full framework also improves both aggregate metrics over direct first-frame conditioning, demonstrating the value of coordinated refinement beyond supplying an initial image. Our main contributions are as follows:
\begingroup
\setlength{\leftmargini}{1.2em}
\setlength{\topsep}{0pt}
\setlength{\partopsep}{0pt}
\begin{itemize}
    \setlength{\itemsep}{6pt}
    \setlength{\parsep}{0pt}
    \setlength{\parskip}{0pt}
    \item We introduce \textbf{VTR-Bench}, a systematic benchmark for evaluating visual text rendering in video generation. By embedding prescribed text in concrete scenes across five application scenarios, VTR-Bench assesses models' ability to render textual information in context, with explicit requirements for its content and carriers.
    \item We develop an automated evaluation pipeline with human alignments that combines carrier-specific transcription and a prompt-specific chain of query to separately assess textual accuracy and fulfillment of video requirements. Our Keyframe-Guided Agentic Framework guides image and video generation through visual evaluation, iterative refinement, and candidate selection.
    \item Experimental results on a wide range of state-of-the-art models reveal systematic difficulties in reproducing textual information across diverse video scenarios, highlighting faithful visual text rendering as an essential capability for advancing video generation.
\end{itemize}
\endgroup

%% file: Sections/related_work.tex
\subsection{Visual Text Rendering in Videos}

Visual text rendering in videos requires preserving textual accuracy under motion, deformation, and changes in visibility. Text-Animator combines text embedding injection, camera control, and text refinement~\citep{liu2024textanimator}. Approaches to this problem span model design and synthetic-data training: HunyuanVideo 1.5 incorporates ByT5-based glyph encoding~\citep{wu2025hunyuanvideo}, while Video Text Preservation fine-tunes Wan2.1 on synthetic text-rich videos~\citep{liu2025videotextpreservation}. Related settings address complementary problems: FlowText synthesizes scene text in existing videos for video text spotting~\citep{zhao2023flowtext}; Dynamic Typography and KineTy animate the glyphs themselves~\citep{liu2024dynamictypography,park2024kinety}; and STRIVE and SteerVTE edit text in source videos~\citep{g2021strive,zeng2026steervte}. VTR-Bench evaluates how accurately video generation models render prescribed text on designated carriers while satisfying the surrounding scene requirements.

\subsection{Video Generation Benchmarks}
Existing benchmarks assess video generation from several complementary perspectives. General benchmarks such as FETV~\citep{liu2023fetv}, VBench~\citep{huang2024vbench}, and EvalCrafter~\citep{liu2024evalcrafter} evaluate generation quality and prompt adherence across multiple dimensions, including visual quality, motion quality, and video-text alignment. VBench-2.0~\citep{zheng2025vbench2} further extends evaluation to intrinsic faithfulness, covering human fidelity, controllability, creativity, physics, and commonsense. Video-Bench~\citep{han2025video} introduces chain-of-query and few-shot scoring to improve alignment with human judgments. Beyond general-purpose evaluation, specialized benchmarks examine specific capabilities: T2V-CompBench~\citep{sun2025t2v} evaluates compositional generation, while TC-Bench~\citep{feng2025tc} examines temporal compositionality. PhyGenBench~\citep{meng2024towards}, VideoPhy~\citep{bansal2025videophy}, and VideoPhy-2~\citep{bansal2026videophy} evaluate physical commonsense, while RulerBench~\citep{he2025ruler} and Sci-VBench~\citep{zhang2026sci} evaluate reasoning capabilities. As video generation models increasingly incorporate audio, recent benchmarks have expanded evaluation to audiovisual generation~\citep{mao2024tavgbench,cao2025t2av,hua2026vabench,liu2026javisdit,yang2026avbench}. Beyond audiovisual synchronization and cross-modal alignment, PhyAVBench~\citep{xie2025phyavbench} and AV-Phys Bench~\citep{cui2026joint} extend the evaluation of physical plausibility to audio-video generation. LongAV-Compass~\citep{liu2026longav}, MSAVBench~\citep{wei2026msavbench}, and MultiRef-Compass~\citep{zhang2026multiref} target minute-scale, multi-shot, and multi-reference-conditioned audio-video generation, respectively. Text rendering has also received dedicated attention. T2VTextBench~\citep{guo2025t2vtextbench} uses human evaluation to assess on-screen text fidelity and temporal consistency, while AVGen-Bench~\citep{zhou2026avgen} incorporates scene text rendering into its broader audiovisual evaluation suite. VTR-Bench evaluates visual text rendering across five application scenarios by automatically transcribing text from designated carriers and computing word error rate against reference text.

%% file: Sections/VTR-Bench.tex
\begin{figure}[t]
    \setlength{\abovecaptionskip}{4pt}
    \centering
    \includegraphics[width=\linewidth]{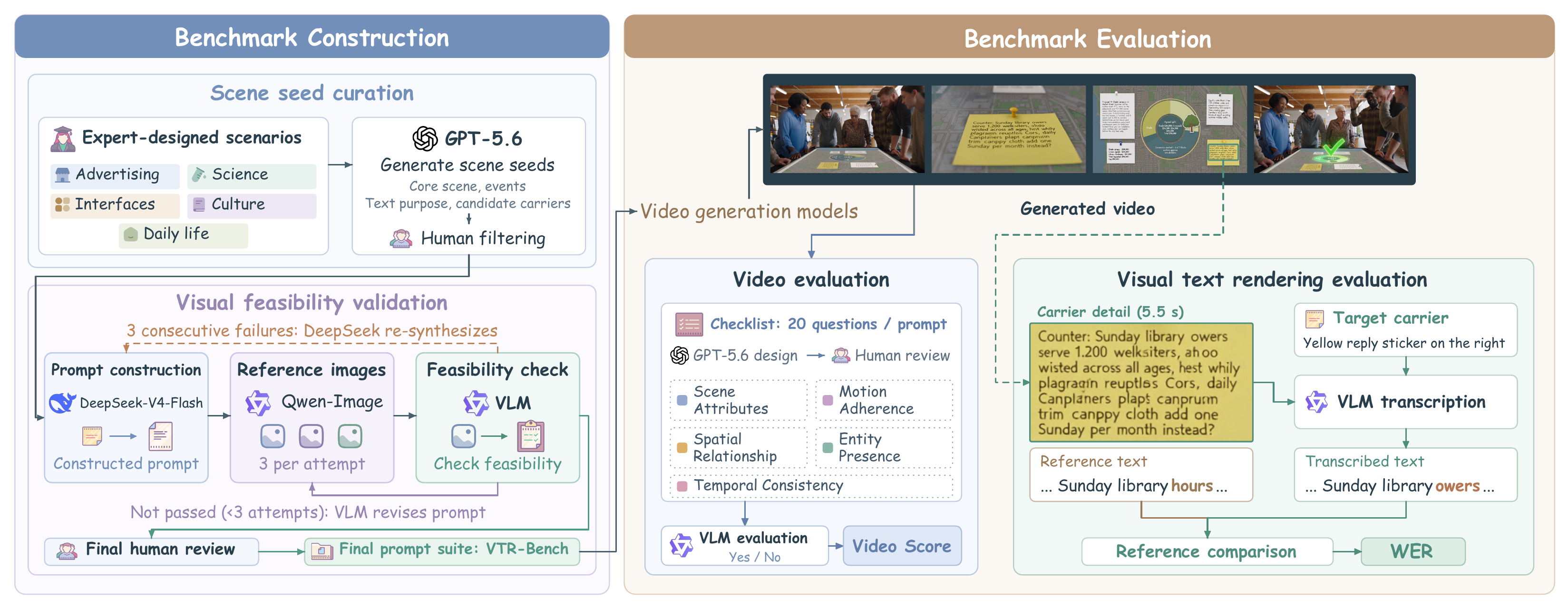}
    \caption{Overview of VTR-Bench. Left: five expert-designed scenario categories guide scene seed generation and human filtering. The retained seeds support prompt construction, followed by iterative refinement through reference image generation and VLM review and a final human review. Right: generated videos are evaluated along two separate dimensions. Video Score measures adherence to scene and video requirements using a 20-question checklist, while WER measures visual text fidelity by comparing carrier-specific VLM transcriptions with reference text.}
    \label{fig:overvieww}
\end{figure}

\subsection{Dataset Construction}

\label{sec:dataset-construction}

We build the prompt suite of VTR-Bench via a multi-stage pipeline that combines scene seed curation, prompt construction, and visual feasibility validation, as shown in Figure~\ref{fig:overvieww}. To establish broad coverage of visual text in context, human experts firstly define five high-level scenario categories. Within these categories, GPT-5.6~\citep{openai2026gpt56} generates diverse scene seeds describing the core scene and events, the purpose of the text, and candidate carriers. These seeds are then screened and deduplicated by human reviewers to balance scenario coverage. Building on the curated seeds, DeepSeek-V4-Flash~\citep{deepseekai2026deepseekv4} is used to construct complete prompts that specify textual content, carrier assignments, and semantic relationships, thereby grounding the texts in concrete scene contexts.

To assess whether the specified text and carriers can be accommodated within a coherent scene, we design an iterative validation mechanism that combines reference images, VLM feedback, and final human review. Specifically, three reference images produced by an image generation model serve as visual evidence for a VLM to assess each candidate prompt and identify requirements that need revision. This feedback guides prompt refinement, with each revised candidate evaluated using three newly generated images. The mechanism also includes a re-synthesis step: three consecutive unsuccessful checks trigger DeepSeek-V4-Flash to reconstruct the candidate before restarting validation. To verify the resulting prompts before inclusion, a final human review follows successful visual validation. The resulting suite contains 300 prompts spanning advertising, science, user interfaces, culture, and daily life. Within this suite, each sample pairs a video generation prompt with reference strings and carrier descriptions, making both the intended text and its location explicit for subsequent evaluation. Further details of dataset construction are provided in Appendix~\ref{app:dataset-details}.

\subsection{Dataset Statistics}
\label{sec:dataset-statistics}

\begin{figure}[t]
    \centering
    \renewcommand{\scriptsize}{\fontsize{6}{7.2}\selectfont}
    \captionsetup[subfloat]{font=scriptsize,justification=centering,singlelinecheck=false,captionskip=2pt,farskip=0pt,nearskip=0pt}
    \begin{minipage}[b]{0.40\linewidth}
        \centering
        \subfloat[Scenario taxonomy\label{fig:statistics-taxonomy}]{\includegraphics[width=\linewidth]{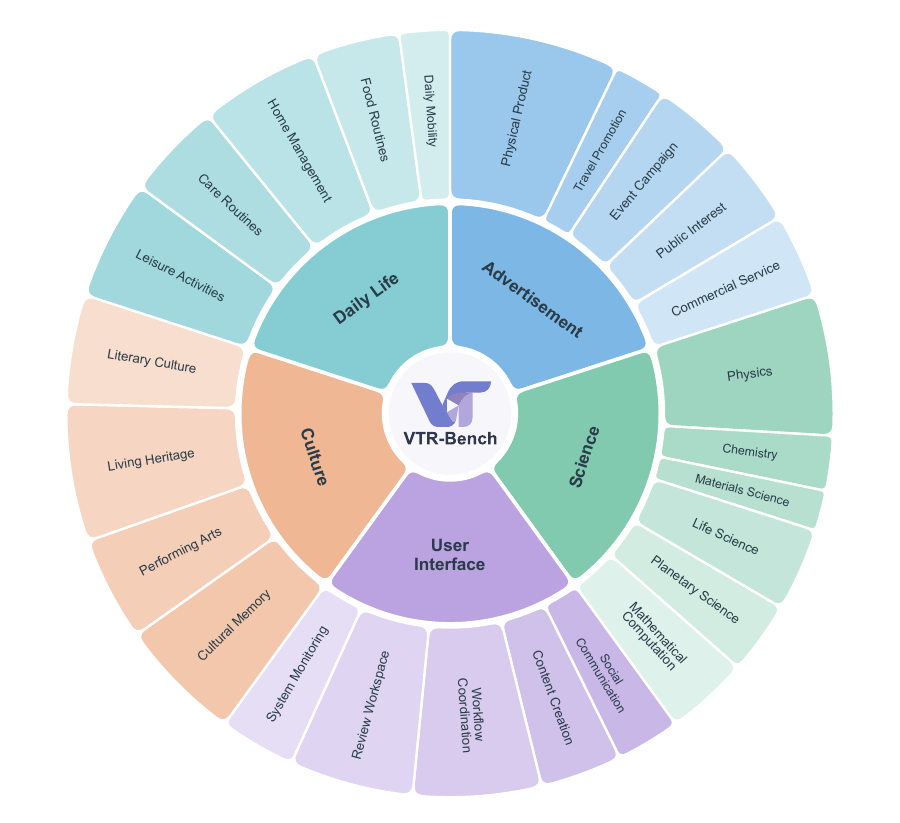}}
    \end{minipage}%
    \begin{minipage}[b]{0.22\linewidth}
        \centering
        \subfloat[\mbox{Distribution of text block counts}\label{fig:statistics-block-counts}]{\includegraphics[width=\linewidth]{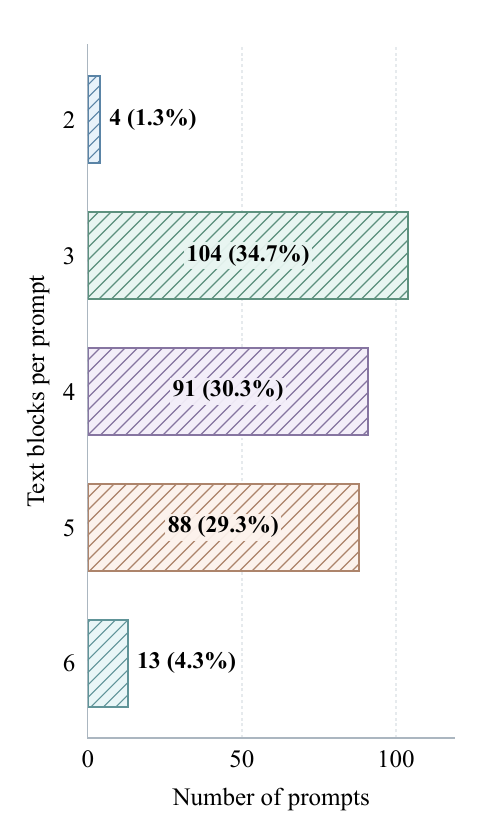}}
    \end{minipage}%
    \begin{minipage}[b]{0.38\linewidth}
        \centering
        \subfloat[\mbox{Distribution of total required text length per video}\label{fig:statistics-video-text-length}]{\includegraphics[width=\linewidth]{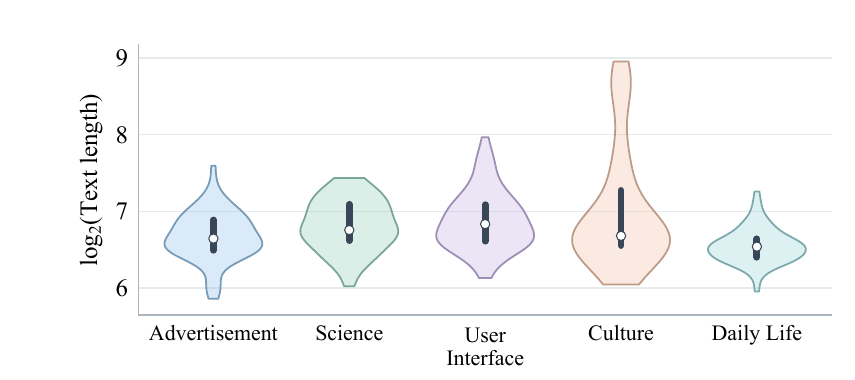}}\par
        \vspace{2pt}
        \subfloat[\mbox{Distribution of individual text block lengths}\label{fig:statistics-block-text-length}]{\includegraphics[width=\linewidth]{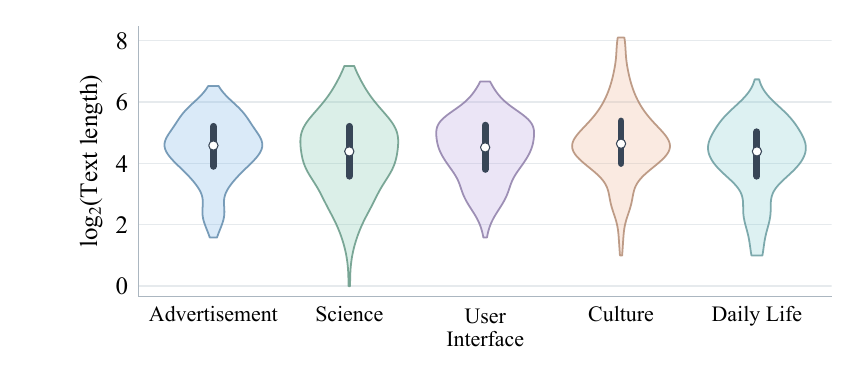}}
    \end{minipage}
    \caption{Dataset statistics of VTR-Bench.}
    \label{fig:dataset-statistics}
    \vspace{-3mm}
\end{figure}

VTR-Bench contains 300 prompts evenly distributed across five application scenarios, with 60 per category. Figure~\ref{fig:statistics-taxonomy} summarizes the 25 sub-scenes covered. Each prompt requires models to render multiple text blocks within a scene, with textual content ranging from short labels to extended passages. A \emph{text block} denotes an annotated textual target with a reference string and a specified carrier. As shown in Figure~\ref{fig:statistics-block-counts}, the suite contains 1,202 text blocks, with two to six blocks per prompt and 94.3\% of prompts requiring three to five blocks. Beyond this multiplicity, the suite spans a broad range of text lengths, measured using the evaluation tokenizer described in Section~\ref{sec:benchmark-evaluation}. The distributions in Figures~\ref{fig:statistics-video-text-length} and~\ref{fig:statistics-block-text-length} show that the total required text length per video ranges from 58 to 496 tokens, with a median of 102.5, while individual blocks range from 1 to 275 tokens, with a median of 23. Together, these properties make VTR-Bench a test of both rendering multiple textual targets within a scene and reproducing longer passages accurately.

\subsection{Benchmark Evaluation}
\label{sec:benchmark-evaluation}

To distinguish fulfillment of scene and motion requirements from visual text accuracy, we design a decoupled evaluation pipeline that measures these two aspects through Video Score and word error rate (WER). Figure~\ref{fig:overvieww} illustrates the two evaluation branches: prompt-specific chain-of-query (CoQ) evaluation and carrier-specific text transcription followed by reference comparison.

\paragraph{Video Evaluation.}
To assess how faithfully a video realizes the requested scene and motion, we adopt query-based evaluation~\citep{han2025video,li2026bizgeneval} and construct a CoQ of 20 questions for each prompt. Generated by GPT-5.6~\citep{openai2026gpt56} and reviewed by human annotators, the queries are tailored to the requirements of each prompt. Across the prompt suite, they span five dimensions: \emph{Scene Attributes}, \emph{Motion Adherence}, \emph{Spatial Relationship}, \emph{Entity Presence}, and \emph{Temporal Consistency}, with each CoQ addressing the dimensions relevant to its prompt. Each question expresses an observable requirement, allowing a VLM to evaluate its fulfillment with a \texttt{yes} or \texttt{no} answer. Video Score is the proportion of satisfied requirements, $\operatorname{VideoScore}(v)=\frac{1}{20}\sum_{j=1}^{20}y_{v,j}$, where $y_{v,j}=1$ for a \texttt{yes} answer and $0$ otherwise.

\paragraph{Visual Text Rendering Evaluation.}
For visual text evaluation, a VLM transcribes each specified carrier from its clearest occurrence in the video, guided by carrier descriptions and the generation prompt with reference text masked. Transcriptions preserve rendering errors and omit unreadable spans; missing or entirely unreadable targets yield empty strings. We then compare these transcriptions with the reference text using WER.

To quantify textual accuracy, we tokenize the transcriptions and reference strings using a shared deterministic tokenizer that preserves case, content-bearing symbols, and individual CJK characters while ignoring ordinary prose punctuation. For target $k$ in video $v$, let $R_{v,k}$ and $H_{v,k}$ denote the reference and hypothesis token sequences. Allowing $\alpha$ additional tokens beyond each reference length, the bounded WER aggregates edit distances across the $K_v$ targets in video $v$:
\begin{equation}
    \operatorname{WER}_{\alpha}(v)=\min\!\left(1,
    \frac{\sum_{k=1}^{K_v} d\!\left(R_{v,k},H_{v,k}[1:|R_{v,k}|+\alpha]\right)}
         {\sum_{k=1}^{K_v}|R_{v,k}|}\right),
\end{equation}
where $H[1:m]$ retains up to the first $m$ tokens and $d$ denotes token-level Levenshtein distance. We set $\alpha = 5$ as our default evaluation setting.

\subsection{Keyframe-Guided Agentic Generation}
\label{sec:agentic-baseline}

\begin{figure}[t]
    \setlength{\abovecaptionskip}{4pt}
    \centering
    \includegraphics[width=\linewidth]{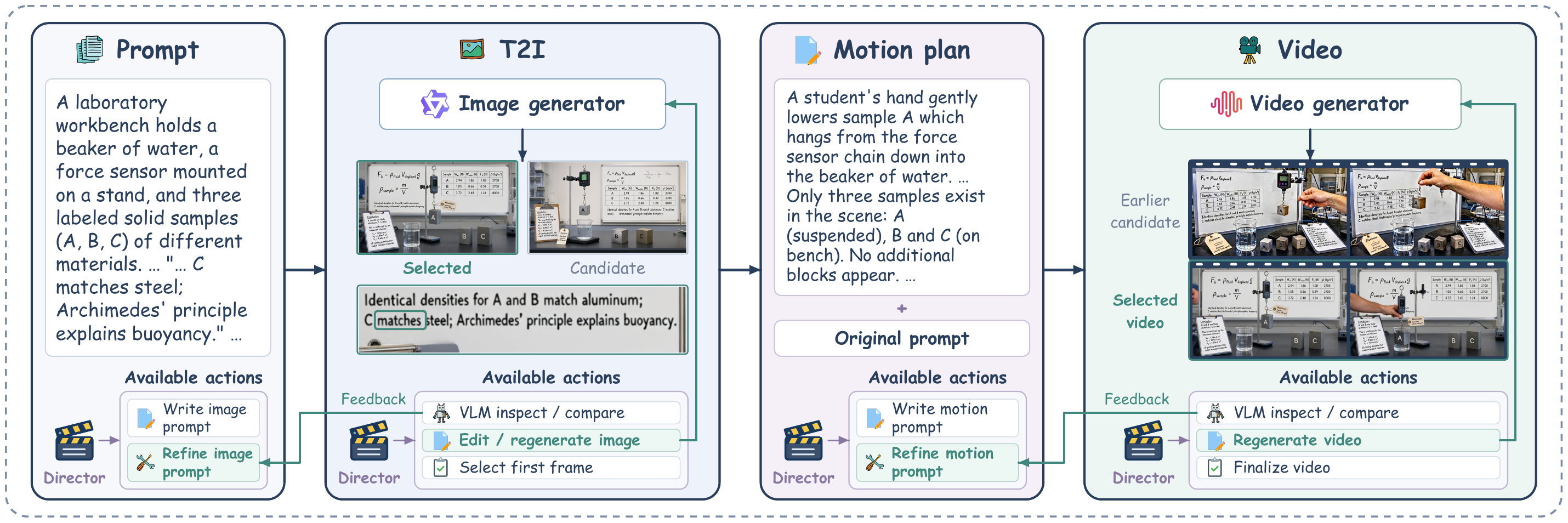}
    \par
    \caption{Keyframe-guided agentic generation. A Director agent uses visual feedback to refine image prompts, edit or regenerate first frames, and revise motion plans. The selected first frame, original prompt, and motion plan condition video generation.}
    \label{fig:agentic-workflow}
    \vspace{-3mm}
\end{figure}

To explore inference-time control of visual text rendering, we design a keyframe-guided agentic generation framework that establishes a first-frame representation of the requested scene before introducing motion. As illustrated in Figure~\ref{fig:agentic-workflow}, a Director agent coordinates image generation, motion planning, and video generation through visual feedback. Given the generation prompt, the Director agent constructs an image prompt and requests candidate first frames. A VLM inspects these candidates for text accuracy, legibility, carrier coverage, and scene consistency. Based on these observations, the Director agent can compare candidates, refine the image prompt for another generation, or edit an existing candidate by supplying the image and a targeted editing instruction. This feedback guides selection of the first frame that will condition video generation.

To animate the selected scene, the Director agent constructs a motion plan specifying subject actions, camera movement, and temporal progression. The video generator receives the selected first frame together with the original prompt and the motion plan, retaining the same model weights used for direct video generation. A VLM then reviews sampled video frames for text stability, carrier persistence, motion coherence, and adherence to the requested content. These observations guide motion-plan refinement and further video generation, while candidate comparison supports final video selection. Throughout the process, the Director agent chooses subsequent actions and their instructions from the available tools based on accumulated visual evidence.

%% file: Sections/experiments.tex
\subsection{Experiment Settings}
\label{sec:experiment-settings}

\paragraph{Evaluation Models.}

We evaluate a variety of models, including both open-sourced models and proprietary models. For open-sourced models, we test Wan2.2-TI2V-5B~\citep{wan2025}, Hunyuanvideo-1.5~\citep{wu2025hunyuanvideo}, LTX-2.3~\citep{hacohen2025ltx2}, Lingbot-Video~\citep{ma2026scaling}, including Lingbot-Video-Dense and Lingbot-Video-MOE and Minimax H3~\citep{minimax2026h3}. For Proprietary Models, we evaluate kling v3.0~\citep{kuaishou2024kling}, happyhorse1.1~\citep{happyhorse2026}, ViduQ3~\citep{vidu2026q3}, Wan-3.0~\citep{wanai2026wan3} and seedance2.5~\citep{seedance202625}.

\paragraph{Configuration.}

For open sourced models, we use the unified framework vllm-omni~\citep{yin2026vllm} for generation, and each model is set with the default optimal generation settings of itself, except for the video length. For Proprietary models, we set the resolution to 720p. The video length is fixed to 10 seconds with a 24 FPS. All of the generation experiments are conducted on NVIDIA H20 GPUs. For VLM evaluator, we use Qwen3.8-27B~\citep{qwen38} for both video and visual text rendering evaluation. Evaluation runs on NVIDIA A800 GPUs.

\subsection{Main Results}

\input{Tables/main_results}

We report the main results in Table~\ref{tab:main_results}, revealing two key findings. \textbf{Visual text rendering remains challenging for current video generation models.} High WERs are prevalent across the evaluated models, and even the strongest model records an overall WER of 0.250. Among open-source models, Minimax H3 stands out with a WER of 0.447, while most others remain close to the metric's upper bound across the five scenarios. This broad pattern shows that their text rendering difficulties extend across application contexts. Proprietary models also differ substantially in text fidelity. Wan3.0 leads text fidelity in every scenario, with Seedance2.5 following within this group, but several other proprietary models still produce substantial text errors and do not match Minimax H3. The results therefore reveal substantial differences within each group, alongside a shared challenge in reproducing the requested text faithfully. VTR-Bench distinguishes these capabilities by testing whether models reproduce the specified textual content across diverse scene contexts.

\textbf{High Video Scores do not guarantee accurate visual text.} Proprietary models achieve consistently high Video Scores, yet differ substantially in text fidelity. Kling v3.0 and Seedance2.5 provide a clear example: both score approximately 0.79 on video requirements, while their WERs are 0.979 and 0.641, respectively. Thus, nearly identical fulfillment of scene and motion requirements can accompany markedly different text accuracy. The same distinction appears across model groups: Minimax H3 renders text more accurately than most proprietary models despite receiving a lower Video Score. These comparisons show that a model's ability to realize the requested scene does not establish whether the text conveys the intended information. Correct actions, entities, and spatial relationships can coexist with incorrect textual content. By evaluating these aspects separately, VTR-Bench exposes a gap that a high Video Score can obscure and identifies visual text rendering as a distinct dimension of video generation capability.

\subsection{Failure Analysis}
\label{sec:failure-analysis}

To better understand visual text rendering failures, we examine them from two perspectives: the effect of generation settings and the transcription outcomes reported by the VLM evaluator.

\begin{figure}[!htb]
    \vspace{-1mm}
    \setlength{\abovecaptionskip}{4pt}
    \centering
    \begin{minipage}[t]{0.49\linewidth}
        \centering
        \includegraphics[width=\linewidth]{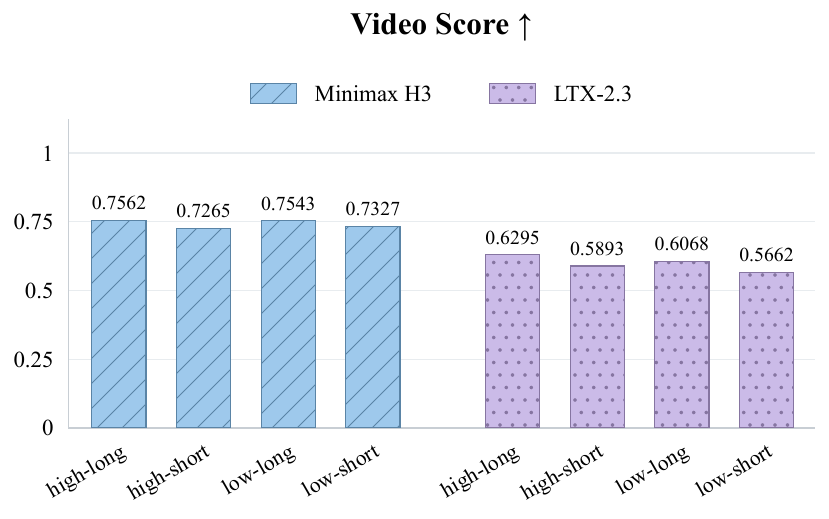}
    \end{minipage}\hfill
    \begin{minipage}[t]{0.49\linewidth}
        \centering
        \includegraphics[width=\linewidth]{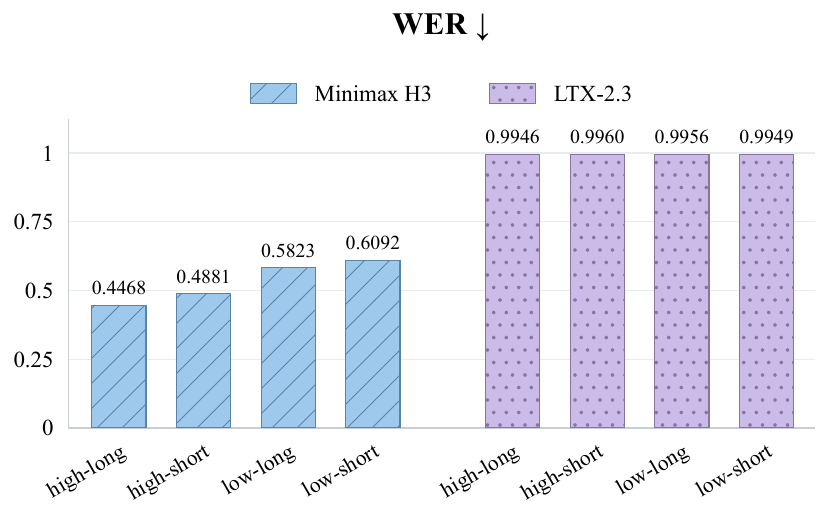}
    \end{minipage}
    \par
    \caption{Effects of generation resolution and duration. Left: Video Score ($\uparrow$). Right: WER ($\downarrow$).}
    \label{fig:resolution-duration}
    \vspace{-1mm}
\end{figure}

\paragraph{Effects of Resolution and Duration.}
Higher resolution can allocate more pixels to text at a comparable relative scale, while shorter videos reduce the temporal span over which text must remain consistent. To examine how these factors affect text fidelity, we compare four combinations of high/low resolution and long/short duration for Minimax H3 and LTX-2.3. For Minimax H3, the high and low resolutions are $1344\times768$ and $832\times480$, respectively; for LTX-2.3, they are $1280\times704$ and $768\times512$. Long and short durations correspond to 10 and 5 seconds for both models.

\textbf{The response to generation settings reflects differences in underlying text-rendering capability.} As shown in Figure~\ref{fig:resolution-duration}, higher resolution reduces WER for Minimax H3 at both durations, allowing its stronger text-rendering ability to benefit from greater spatial detail. In contrast, LTX-2.3 maintains a WER close to 1 across all four configurations despite changes in Video Score. Shortening duration likewise provides no consistent improvement in text fidelity across the two models. Together, these results suggest that higher resolution can enhance an existing capacity to render text, while configuration changes alone do not resolve the text errors observed in LTX-2.3. Strengthening the model's underlying text-rendering capability is therefore central to realizing the benefits of improved generation settings.

\begin{wrapfigure}[18]{r}{0.54\textwidth}
    \vspace{-\intextsep}
    \centering
    \includegraphics[width=\linewidth]{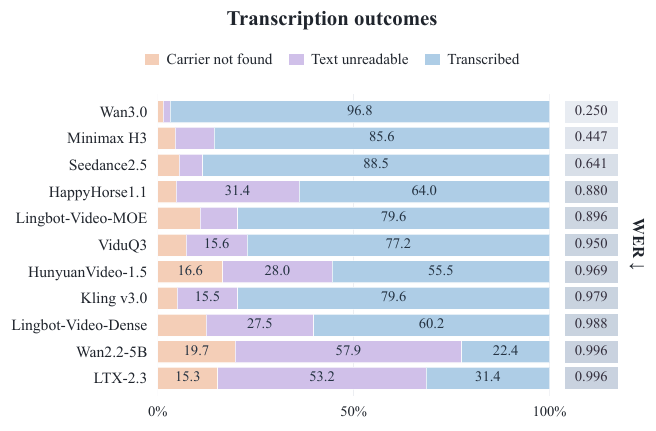}
    \par
    \setlength{\abovecaptionskip}{4pt}
    \caption{Transcription outcomes reported by VLM evaluator. Bars show the proportion of text targets in each state.}
    \label{fig:readability-outcomes}
    \vspace{-1mm}
\end{wrapfigure}
\paragraph{Transcription Outcomes.}
To examine how these capability differences manifest in generated text, we analyze Qwen3.8-27B's transcription outcomes for 13,213 text targets across videos. The evaluator uses three states predefined before evaluation: \emph{carrier not found}, \emph{text unreadable}, and \emph{transcribed}. Here, \emph{text unreadable} indicates that the carrier is found but its entire text is unreadable. \mbox{Figure~\ref{fig:readability-outcomes}} shows the distribution of these states. Across models, the evaluator reports unreadable text for 23.31\% of targets, compared with 9.34\% for carriers not found. This pattern points to difficulties recovering text even after its carrier has been located. Such outcomes are particularly prevalent for Wan2.2-5B and LTX-2.3, where over half of the targets are judged unreadable, consistent with their high WERs.

\textbf{Transcribability does not establish textual accuracy.} Kling v3.0 has a transcription rate of 79.62\% but an overall WER of 0.979, whereas Minimax H3 combines a transcription rate of 85.61\% with a WER of 0.447. Thus, substantial transcription coverage can coexist with extensive errors against the required content. These observations identify two priorities for visual text rendering: legible text on the intended carriers and accurate reproduction of the prescribed content.

\subsection{Effect of Test-Time Refinement}
\label{sec:test-time-refinement}

To examine whether inference-time control improves visual text rendering, we compare three generation settings using Minimax H3: direct generation from the prompt, denoted as \emph{Original}; \emph{I2V}, which takes the original prompt and the first image generated by the agentic framework as input; and \emph{Agentic}, the full framework described in Section~\ref{sec:agentic-baseline}. For the agentic framework, Qwen3.7-plus~\citep{qwen37} serves as both the Director agent and the VLM, while Qwen-Image-3.0~\citep{qwenimage30} serve as the image generator. For both I2V and Agentic generation, we remove the first frame of each generated video before evaluation.

\begin{figure}[!htb]
    \vspace{-1mm}
    \centering
    \includegraphics[width=\linewidth]{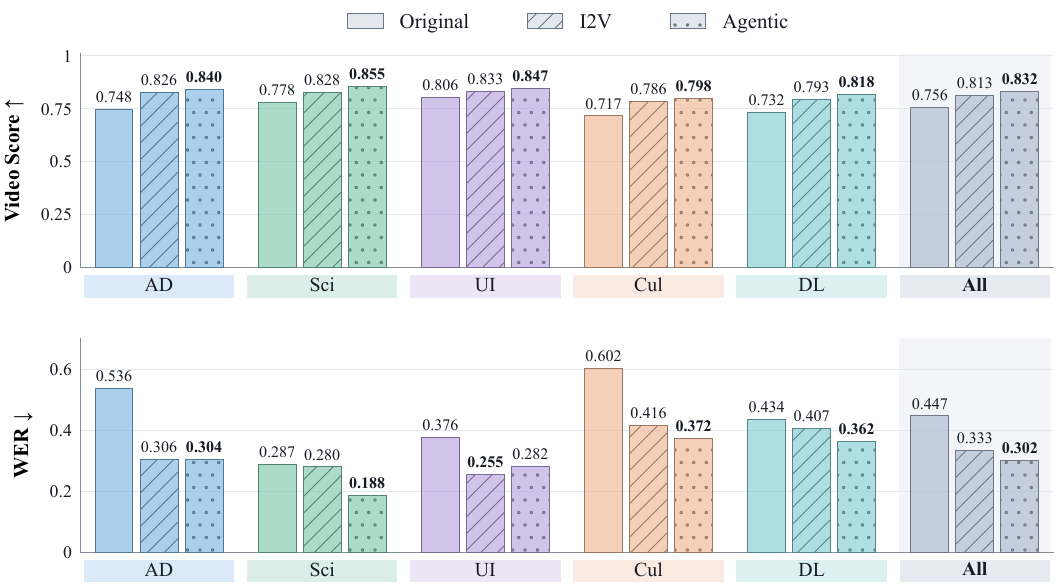}
    \par
    \setlength{\abovecaptionskip}{4pt}
    \caption{Comparison of direct generation, I2V, and the agentic framework on Minimax H3. Bold values indicate the best result within each scenario or overall.}
    \label{fig:agentic-comparison}
    \vspace{-1mm}
\end{figure}

\textbf{Test-time refinement improves both text fidelity and fulfillment of video requirements.} As shown in Figure~\ref{fig:agentic-comparison}, the agentic framework reduces overall WER from 0.4468 to 0.3015, a 32.5\% relative reduction, while raising Video Score from 0.7562 to 0.8315. Both metrics improve across all five scenarios relative to direct generation. These results show that explicit first-frame construction and visual feedback can improve text fidelity alongside adherence to scene and motion requirements, using the same video generator.

\textbf{The full framework provides gains beyond first-frame conditioning alone.} I2V improves both aggregate metrics over direct generation, while the full framework further reduces overall WER by 9.4\% relative to I2V and achieves the highest Video Score in every scenario. Its WER advantage over I2V spans four of the five scenarios and is largest in scientific videos; in UI scenes, I2V achieves the lowest WER, while the full framework achieves the highest Video Score. Thus, the full framework improves on initial image conditioning through visual refinement and motion planning.

\subsection{Human Alignment}
\label{sec:human-alignment}

To assess alignment with human judgments, we compare Qwen3.8-27B's evaluations with human annotations on 100 sampled videos, covering 2,000 CoQ judgments and 404 text blocks. Guided by the transcription outcomes in Figure~\ref{fig:readability-outcomes}, we sample more videos from Seedance2.5, Minimax H3, and Wan3.0, which combine high transcription rates with relatively low overall WERs. This choice prioritizes examples with recoverable text for human comparison. Three human annotators provide annotations for each binary query and text block. For each binary query, the majority vote serves as the human reference. As shown in \mbox{Table~\ref{tab:human-alignment}}, the automated evaluator agrees with this majority on 92.15\% of the queries, indicating consistent judgments of scene and motion requirement fulfillment.

\begin{wraptable}{r}{0.50\textwidth}
    \vspace{-\intextsep}
    \setlength{\abovecaptionskip}{0pt}
    \centering
    \caption{Alignment of Qwen3.8-27B.}
    \label{tab:human-alignment}
    \footnotesize
    \setlength{\tabcolsep}{2pt}
    \renewcommand{\arraystretch}{1.18}
    \resizebox{\linewidth}{!}{%
    \begin{tabular}{@{}lccccc@{}}
        \toprule
        Evaluation & Agreement $\uparrow$ & Pearson $\uparrow$ & Spearman $\uparrow$ & Lin's CCC $\uparrow$ & MAE $\downarrow$ \\
        \midrule
        CoQ judgments & 92.15\% & --- & --- & --- & --- \\
        Text-block WER & --- & 0.9541 & 0.9370 & 0.9536 & 0.0545 \\
        Video-level WER & --- & 0.9842 & 0.9731 & 0.9835 & 0.0407 \\
        \bottomrule
    \end{tabular}%
    }
\end{wraptable}

For visual text evaluation, we compute WER from each transcription against the same benchmark reference text. We average the three annotators' WERs for each text block or video and compare these averages with the corresponding VLM-derived scores. The strong Pearson and Spearman correlations indicate close agreement between human and VLM assessments of text error severity and relative rankings. This correspondence also extends to numerical agreement: video-level Lin's concordance correlation coefficient reaches 0.9835, with a mean absolute error of 0.0407. Together, these results demonstrate strong agreement between automated and human evaluation across both branches.

%% file: Tables/main_results.tex
\begin{table*}[!t]
\centering
\caption{Main results on VTextBench. The top three results in each metric are highlighted in blue, with darker shades indicating better performance.}
{\small
\definecolor{resultblue}{HTML}{7390BA}
\newcommand{\rankfirst}[1]{\cellcolor{resultblue!65}#1}
\newcommand{\ranksecond}[1]{\cellcolor{resultblue!40}#1}
\newcommand{\rankthird}[1]{\cellcolor{resultblue!20}#1}
\setlength{\tabcolsep}{3pt}
\renewcommand{\arraystretch}{1.10}
\resizebox{\textwidth}{!}{%
\begin{tabular}{@{}l|r@{\,}c@{\,}l|*{6}{c}|*{6}{c}@{}}
\toprule
\multicolumn{1}{c|}{\multirow{2}{*}{\textbf{Model}}} &
\multicolumn{3}{c|}{\multirow{2}{*}{\textbf{Resolution}}} &
\multicolumn{6}{c|}{\textbf{WER} ($\downarrow$)} &
\multicolumn{6}{c}{\textbf{Video Score} ($\uparrow$)} \\
\cmidrule(lr){5-10}
\cmidrule(l){11-16}
& \multicolumn{3}{c|}{} &
\textbf{AD} & \textbf{Sci} & \textbf{UI} &
\textbf{Cul} & \textbf{DL} & \textbf{All} &
\textbf{AD} & \textbf{Sci} & \textbf{UI} &
\textbf{Cul} & \textbf{DL} & \textbf{All} \\
\midrule
\rowcolor{black!7}
\multicolumn{16}{l}{\textbf{Open-Sourced Models}} \\
\addlinespace[1.5pt]
Wan2.2-5B        & 1280 & $\times$ & 704 & 0.994 & 0.993 & 0.999 & 0.998 & 0.994 & 0.996 & 0.523 & 0.475 & 0.386 & 0.357 & 0.438 & 0.436 \\
HunyuanVideo-1.5 & 1280 & $\times$ & 720 & 0.972 & 0.952 & 0.974 & 0.985 & 0.964 & 0.969 & 0.606 & 0.628 & 0.653 & 0.557 & 0.562 & 0.601 \\
LTX-2.3          & 768  & $\times$ & 512 & 0.997 & 0.992 & 0.998 & 0.998 & 0.993 & 0.996 & 0.682 & 0.632 & 0.569 & 0.560 & 0.590 & 0.607 \\
Lingbot-Video-Dense & 832 & $\times$ & 480 & 0.992 & 0.974 & 0.993 & 0.996 & 0.988 & 0.988 & 0.573 & 0.571 & 0.556 & 0.469 & 0.503 & 0.534 \\
Lingbot-Video-MOE & 832 & $\times$ & 480 & 0.895 & 0.841 & 0.900 & 0.927 & 0.916 & 0.896 & 0.553 & 0.621 & 0.599 & 0.492 & 0.480 & 0.549 \\
Minimax H3       & 1344 & $\times$ & 768 & \ranksecond{0.536} & \ranksecond{0.287} & \ranksecond{0.376} & \ranksecond{0.602} & \ranksecond{0.434} & \ranksecond{0.447} & 0.748 & 0.778 & 0.806 & 0.717 & 0.732 & 0.756 \\

\midrule
\rowcolor{black!7}
\multicolumn{16}{l}{\textbf{Proprietary Models}} \\
\addlinespace[1.5pt]
ViduQ3       & 1280 & $\times$ & 720 & 0.939 & 0.928 & 0.961 & 0.983 & 0.939 & 0.950 & 0.818 & \rankthird{0.824} & 0.789 & 0.733 & 0.764 & 0.786 \\
HappyHorse1.1    & 1280 & $\times$ & 720 & 0.900 & 0.843 & 0.841 & 0.929 & 0.886 & 0.880 & \ranksecond{0.862} & \ranksecond{0.844} & \ranksecond{0.826} & \ranksecond{0.792} & \ranksecond{0.830} & \ranksecond{0.831} \\
Kling v3.0       & 1280 & $\times$ & 720 & 0.986 & 0.964 & 0.994 & 0.987 & 0.964 & 0.979 & \rankthird{0.820} & 0.798 & 0.794 & \rankthird{0.757} & \rankthird{0.795} & \rankthird{0.793} \\
Seedance2.5      & 1280 & $\times$ & 720 & \rankthird{0.544} & \rankthird{0.627} & \rankthird{0.664} & \rankthird{0.742} & \rankthird{0.627} & \rankthird{0.641} & 0.793 & 0.818 & \rankthird{0.823} & 0.731 & 0.790 & 0.791 \\
Wan3.0           & 1280 & $\times$ & 720 & \rankfirst{0.234} & \rankfirst{0.206} & \rankfirst{0.215} & \rankfirst{0.340} & \rankfirst{0.255} & \rankfirst{0.250} & \rankfirst{0.873} & \rankfirst{0.871} & \rankfirst{0.841} & \rankfirst{0.816} & \rankfirst{0.846} & \rankfirst{0.849} \\
\bottomrule
\end{tabular}%
}
}
\label{tab:main_results}
\end{table*}

%% file: Sections/conclusion.tex
\section{Conclusion}
\label{sec:conclusion}

VTR-Bench establishes visual text rendering as a distinct criterion for assessing whether generated videos faithfully convey the information specified by users. By grounding textual requirements in concrete application scenarios, the benchmark connects text accuracy to the activities, objects, and interfaces through which written information serves a purpose. Our evaluation of 11 state-of-the-art models shows that this capability remains a shared challenge across open-source and proprietary systems, with the strongest model recording an overall WER of 0.250. The resulting failure analysis makes this challenge more concrete: text can fail to become readable, or remain legible while conveying incorrect content. Progress therefore requires both recognizable characters and faithful reproduction of the intended information. The improvements achieved by our Keyframe-Guided Agentic Framework further show that visual feedback can guide generation toward this goal, reducing overall WER by 32.5\% while improving Video Score on Minimax H3. These findings motivate evaluating visual text as part of a video's communicative function, alongside its depiction of scenes and actions. This perspective provides a basis for studying how scene complexity, textual density, and motion affect text fidelity, with explicit content requirements serving as evaluation targets. We hope VTR-Bench will support the development of video generation models whose written content is as faithful to user intent as the scenes they create.

%% file: Sections/Appendix/more_related_work.tex
\section{More Related Work}

\paragraph{Visual text rendering in images.}
Research on image text rendering has developed from explicit character and layout control toward native text modeling and targeted refinement. TextDiffuser, Glyph-ByT5, and AnyText2 establish useful mechanisms for layout planning, character-aware representation, and attribute control~\citep{chen2023textdiffuser,liu2024glyphbyt5,tuo2024anytext2}. Recent work strengthens these capabilities through complementary approaches. Qwen-Image combines text-focused data curation with progressive training that extends to paragraph-level text~\citep{wu2025qwenimage}. TextGuider addresses text omission through attention alignment, while DCText separates long or multiple strings into designated regions using scheduled attention masks~\citep{baek2025textguider,song2025dctext}. GlyphAnchor introduces position-anchored glyph priors, and TextPecker improves structural fidelity through rewards sensitive to character-level anomalies~\citep{xiang2026glyphanchor,zhu2026textpecker}. PosterMaker and BizGen further study product posters and article-level infographics~\citep{gao2025postermaker,peng2025bizgen}. These developments emphasize text completeness, placement, and dense content rendering alongside character accuracy.

%% file: Sections/Appendix/benchmark.tex
\section{Dataset Details}
\label{app:dataset-details}

\subsection{Scene-Grounded Prompt Construction}

To embed visual text in meaningful scene contexts, we use category-specific construction templates that connect the intended writing to the depicted activity. DeepSeek-V4-Flash~\citep{deepseekai2026deepseekv4} constructs prompts from human-screened scene seeds, specifying the text, its physical or interface carriers, and its semantic role. Shared constraints ensure that the writing serves a clear purpose and remains consistent with the scene, while category-specific requirements capture its different uses across the five application scenarios. The construction template is provided below.

\begingroup
\raggedbottom
\tcbset{datasetpromptstyle/.style={breakable,colback=white,colframe=black!25,
    boxrule=0.4pt,arc=1mm,left=5pt,right=5pt,top=5pt,bottom=5pt,
    colbacktitle=black!5,coltitle=black,fonttitle=\small\bfseries,
    fontupper=\footnotesize\ttfamily\fontencoding{T1}\linespread{0.95}\selectfont\raggedright}}
\newcommand{\datasetprompt}[3][]{%
    \begin{tcolorbox}[datasetpromptstyle,title={#2},title after break={#2 (continued)},#1]
        \VerbatimInput[fontfamily=pcr,formatcom=\fontencoding{T1}\selectfont,fontsize=\footnotesize,breaklines=true,breakanywhere=false,
            breaksymbolleft={},breaksymbolright={},baselinestretch=0.95]{#3}
    \end{tcolorbox}%
}

\datasetprompt{Prompt Construction: System Prompt}{Sections/Appendix/prompts/construction_system_refined.txt}

\datasetprompt{Prompt Construction: Output Format}{Sections/Appendix/prompts/construction_output_format.json}

\subsection{Visual Feasibility Validation}
\label{app:visual-validation}

Visual feasibility validation assesses whether the requested text can be coherently accommodated within the scene. For each candidate, three reference images generated by Qwen-Image~\citep{wu2025qwenimage} are jointly reviewed by Qwen3.8-27B~\citep{qwen38} for carrier coverage, text hierarchy, and integration with the depicted activity. Revisions address structural problems that recur in at least two images and are attributable to the prompt, preserving the intended text and its semantic role. This separates prompt feasibility from the generator's text-rendering accuracy. Validated prompts undergo final human review before inclusion in VTR-Bench. The review template is provided below.

\datasetprompt{Visual Feasibility Validation: System Prompt}{Sections/Appendix/prompts/visual_review_system_refined.txt}

\datasetprompt{Visual Feasibility Validation: Output Format}{Sections/Appendix/prompts/visual_review_output_format.txt}

\endgroup

%% file: Sections/Appendix/evaluator_results.tex
\raggedbottom
\makeatletter
\setlength{\@fptop}{0pt}
\makeatother
\section{Additional Evaluation Results}
\label{app:additional-evaluation}

These analyses use the main-experiment videos and the evaluation protocol in Section~\ref{sec:benchmark-evaluation}, with $\alpha=5$ except in the parameter sweep.

\subsection{Results with an Alternative Evaluator}

Table~\ref{tab:qwen36-results} reports the results obtained with Qwen3.6-27B~\citep{qwen3.6-27b}, allowing us to examine whether evaluator choice changes the model comparison.

\input{Tables/qwen36_results}

Both evaluators produce identical overall rankings for WER and Video Score across all 11 models, giving a Spearman correlation of $1.000$ for each metric. Wan3.0, Minimax H3, and Seedance2.5 occupy the top three positions in text fidelity. Qwen3.6-27B assigns higher Video Scores throughout, but the largest difference in overall WER is 0.017. Evaluator choice therefore affects score calibration while preserving the ordering of model performance.

\subsection{Human Alignments and Evaluator Selection}
\label{app:evaluator-human-alignments}

To choose between the evaluators, we extend the human comparison in Section~\ref{sec:human-alignment} to Qwen3.6-27B using the same 100 annotated videos. Table~\ref{tab:evaluator-human-alignments} reports agreement on CoQ judgments and on text-block and video-level WER.

\input{Tables/evaluator_human_alignments}

Qwen3.8-27B agrees with human CoQ judgments on 92.15\% of queries, compared with 91.60\% for Qwen3.6-27B, and performs better on all four text-block alignment metrics. At the video level, it also achieves higher Pearson correlation and Lin's CCC, while Qwen3.6-27B has higher Spearman correlation and lower MAE. We select Qwen3.8-27B for the main evaluation based on its stronger agreement on individual queries and text blocks, the units on which the evaluation pipeline operates.

\input{Sections/Appendix/alpha_sensitivity}

\input{Sections/Appendix/video_dimensions}

\input{Sections/Appendix/failure_analysis}

%% file: Tables/qwen36_results.tex
\begin{table}[!htbp]
\centering
\setlength{\abovecaptionskip}{4pt}
\caption{Results on VTR-Bench evaluated by Qwen3.6-27B. Darker blue highlights indicate better results among the top three models in each score column. Ranks use overall WER and Video Score before rounding.}
\label{tab:qwen36-results}
\begingroup
\definecolor{evaluatorblue}{HTML}{7390BA}
\newcommand{\evalfirst}[1]{\cellcolor{evaluatorblue!65}#1}
\newcommand{\evalsecond}[1]{\cellcolor{evaluatorblue!40}#1}
\newcommand{\evalthird}[1]{\cellcolor{evaluatorblue!20}#1}
\small
\setlength{\tabcolsep}{3pt}
\renewcommand{\arraystretch}{1.12}
\resizebox{\linewidth}{!}{%
\begin{tabular}{@{}l|*{6}{c}|*{6}{c}|cc|cc@{}}
\toprule
\multirow{2}{*}{\textbf{Model}} & \multicolumn{6}{c|}{\textbf{WER} ($\downarrow$)} & \multicolumn{6}{c|}{\textbf{Video Score} ($\uparrow$)} & \multicolumn{2}{c|}{\textbf{WER Rank} ($\downarrow$)} & \multicolumn{2}{c}{\textbf{Video Rank} ($\downarrow$)} \\
\cmidrule(lr){2-7}\cmidrule(lr){8-13}\cmidrule(lr){14-15}\cmidrule(l){16-17}
& \textbf{AD} & \textbf{Sci} & \textbf{UI} & \textbf{Cul} & \textbf{DL} & \textbf{All}
& \textbf{AD} & \textbf{Sci} & \textbf{UI} & \textbf{Cul} & \textbf{DL} & \textbf{All} & \textbf{Qwen3.6} & \textbf{Qwen3.8} & \textbf{Qwen3.6} & \textbf{Qwen3.8} \\
\midrule
\rowcolor{black!7}\multicolumn{17}{l}{\textbf{Open-Sourced Models}} \\
Wan2.2-5B & 0.996 & 0.990 & 0.997 & 0.997 & 0.986 & 0.993 & 0.563 & 0.577 & 0.477 & 0.421 & 0.495 & 0.507 & 11 & 11 & 11 & 11 \\
HunyuanVideo-1.5 & 0.966 & 0.944 & 0.966 & 0.982 & 0.964 & 0.964 & 0.668 & 0.718 & 0.708 & 0.618 & 0.602 & 0.663 & 7 & 7 & 8 & 8 \\
LTX-2.3 & 0.993 & 0.987 & 0.996 & 0.997 & 0.986 & 0.992 & 0.728 & 0.733 & 0.696 & 0.639 & 0.637 & 0.687 & 10 & 10 & 7 & 7 \\
Lingbot-Video-Dense & 0.991 & 0.964 & 0.986 & 0.995 & 0.981 & 0.983 & 0.614 & 0.655 & 0.647 & 0.540 & 0.548 & 0.601 & 9 & 9 & 10 & 10 \\
Lingbot-Video-MOE & 0.876 & 0.837 & 0.891 & 0.916 & 0.888 & 0.881 & 0.654 & 0.680 & 0.687 & 0.573 & 0.550 & 0.629 & 5 & 5 & 9 & 9 \\
Minimax H3 & \evalthird{0.525} & \evalsecond{0.316} & \evalsecond{0.367} & \evalsecond{0.610} & \evalsecond{0.434} & \evalsecond{0.450} & 0.816 & 0.833 & 0.840 & 0.793 & 0.763 & 0.809 & 2 & 2 & 6 & 6 \\
\midrule
\rowcolor{black!7}\multicolumn{17}{l}{\textbf{Proprietary Models}} \\
ViduQ3 & 0.939 & 0.923 & 0.960 & 0.978 & 0.933 & 0.947 & \evalthird{0.848} & 0.855 & 0.838 & 0.792 & 0.823 & 0.831 & 6 & 6 & 5 & 5 \\
HappyHorse1.1 & 0.879 & 0.797 & 0.828 & 0.926 & 0.882 & 0.863 & \evalfirst{0.892} & \evalsecond{0.886} & \evalsecond{0.872} & \evalsecond{0.835} & \evalsecond{0.860} & \evalsecond{0.869} & 4 & 4 & 2 & 2 \\
Kling v3.0 & 0.981 & 0.963 & 0.989 & 0.986 & 0.964 & 0.977 & 0.842 & 0.853 & 0.837 & \evalthird{0.809} & \evalthird{0.844} & \evalthird{0.837} & 8 & 8 & 3 & 3 \\
Seedance2.5 & \evalsecond{0.525} & \evalthird{0.636} & \evalthird{0.653} & \evalthird{0.733} & \evalthird{0.612} & \evalthird{0.632} & 0.835 & \evalthird{0.864} & \evalthird{0.857} & 0.793 & 0.832 & 0.836 & 3 & 3 & 4 & 4 \\
Wan3.0 & \evalfirst{0.241} & \evalfirst{0.203} & \evalfirst{0.203} & \evalfirst{0.361} & \evalfirst{0.234} & \evalfirst{0.248} & \evalfirst{0.892} & \evalfirst{0.908} & \evalfirst{0.879} & \evalfirst{0.878} & \evalfirst{0.873} & \evalfirst{0.886} & 1 & 1 & 1 & 1 \\
\bottomrule
\end{tabular}%
}
\endgroup
\end{table}

%% file: Tables/evaluator_human_alignments.tex
\begin{table}[!t]
\centering
\begin{minipage}{0.72\linewidth}
\setlength{\abovecaptionskip}{4pt}
\caption{Human alignments of the two evaluators. Better results are shown in bold.}
\label{tab:evaluator-human-alignments}
\centering
\small
\setlength{\tabcolsep}{5pt}
\renewcommand{\arraystretch}{1.12}
\begin{tabular*}{\linewidth}{@{}l@{\extracolsep{\fill}}cc@{}}
\toprule
\textbf{Metric} & \textbf{Qwen3.6-27B} & \textbf{Qwen3.8-27B} \\
\midrule
\rowcolor{black!5}\multicolumn{3}{l}{CoQ judgments: 2,000 queries} \\
Agreement $\uparrow$ & 91.60\% & \textbf{92.15\%} \\
\midrule
\rowcolor{black!5}\multicolumn{3}{l}{Text-block WER: 404 blocks} \\
Pearson $\uparrow$ & 0.9536 & \textbf{0.9541} \\
Spearman $\uparrow$ & 0.9365 & \textbf{0.9370} \\
Lin's CCC $\uparrow$ & 0.9527 & \textbf{0.9536} \\
MAE $\downarrow$ & 0.0582 & \textbf{0.0545} \\
\midrule
\rowcolor{black!5}\multicolumn{3}{l}{Video-level WER: 100 videos} \\
Pearson $\uparrow$ & 0.9793 & \textbf{0.9842} \\
Spearman $\uparrow$ & \textbf{0.9759} & 0.9731 \\
Lin's CCC $\uparrow$ & 0.9790 & \textbf{0.9835} \\
MAE $\downarrow$ & \textbf{0.0398} & 0.0407 \\
\bottomrule
\end{tabular*}
\end{minipage}
\end{table}

%% file: Sections/Appendix/alpha_sensitivity.tex
\subsection{Sensitivity to the WER Truncation Parameter}
\label{app:alpha-sensitivity}

We recompute WER from the saved transcriptions for $\alpha\in\{0,\ldots,10\}$, varying how many tokens beyond the reference length are retained. Table~\ref{tab:alpha-sensitivity} shows the resulting scores for both evaluators, with the default setting highlighted.

\input{Tables/alpha_sensitivity}

\textbf{The three leading models retain their ordering across all truncation settings.} Wan3.0, Minimax H3, and Seedance2.5 achieve the lowest overall WERs, in that order, at every value of $\alpha$ under both evaluators. Their relative performance is therefore stable as the allowance for additional transcription tokens changes.

%% file: Tables/alpha_sensitivity.tex
\begin{table}[!t]
\centering
\setlength{\abovecaptionskip}{4pt}
\caption{Overall WER for $\alpha=0$--$10$ under both evaluators. The shaded column marks the default $\alpha=5$; bold values indicate the lowest WER in each column within each evaluator.}
\label{tab:alpha-sensitivity}
\begingroup
\definecolor{alphablue}{HTML}{7390BA}
\small
\setlength{\tabcolsep}{3.5pt}
\renewcommand{\arraystretch}{1.10}
\resizebox{\linewidth}{!}{%
\begin{tabular}{@{}l|*{11}{c}@{}}
\toprule
\multirow{2}{*}{\textbf{Model}} & \multicolumn{11}{c}{\textbf{WER} ($\downarrow$)} \\
\cmidrule(l){2-12}
& $\alpha=0$ & $\alpha=1$ & $\alpha=2$ & $\alpha=3$ & $\alpha=4$ & \cellcolor{alphablue!20}$\alpha=5$ & $\alpha=6$ & $\alpha=7$ & $\alpha=8$ & $\alpha=9$ & $\alpha=10$ \\
\midrule
\rowcolor{black!7}\multicolumn{12}{l}{\textbf{Qwen3.8-27B}} \\
Wan2.2-5B & 0.993 & 0.994 & 0.995 & 0.995 & 0.996 & \cellcolor{alphablue!20}0.996 & 0.996 & 0.996 & 0.996 & 0.996 & 0.996 \\
HunyuanVideo-1.5 & 0.955 & 0.959 & 0.963 & 0.966 & 0.968 & \cellcolor{alphablue!20}0.969 & 0.971 & 0.972 & 0.973 & 0.973 & 0.974 \\
LTX-2.3 & 0.991 & 0.993 & 0.994 & 0.995 & 0.995 & \cellcolor{alphablue!20}0.996 & 0.996 & 0.996 & 0.996 & 0.996 & 0.996 \\
Lingbot-Video-Dense & 0.971 & 0.978 & 0.982 & 0.986 & 0.987 & \cellcolor{alphablue!20}0.988 & 0.989 & 0.990 & 0.990 & 0.991 & 0.991 \\
Lingbot-Video-MOE & 0.858 & 0.866 & 0.875 & 0.882 & 0.889 & \cellcolor{alphablue!20}0.896 & 0.901 & 0.907 & 0.911 & 0.915 & 0.918 \\
Minimax H3 & 0.444 & 0.443 & 0.443 & 0.443 & 0.445 & \cellcolor{alphablue!20}0.447 & 0.449 & 0.451 & 0.453 & 0.455 & 0.457 \\
ViduQ3 & 0.933 & 0.938 & 0.942 & 0.946 & 0.948 & \cellcolor{alphablue!20}0.950 & 0.951 & 0.953 & 0.954 & 0.954 & 0.955 \\
HappyHorse1.1 & 0.862 & 0.866 & 0.870 & 0.874 & 0.877 & \cellcolor{alphablue!20}0.880 & 0.883 & 0.885 & 0.887 & 0.889 & 0.890 \\
Kling v3.0 & 0.949 & 0.959 & 0.967 & 0.973 & 0.976 & \cellcolor{alphablue!20}0.979 & 0.981 & 0.983 & 0.984 & 0.984 & 0.985 \\
Seedance2.5 & 0.631 & 0.632 & 0.634 & 0.636 & 0.638 & \cellcolor{alphablue!20}0.641 & 0.643 & 0.646 & 0.649 & 0.652 & 0.654 \\
Wan3.0 & \textbf{0.255} & \textbf{0.252} & \textbf{0.250} & \textbf{0.249} & \textbf{0.250} & \cellcolor{alphablue!20}\textbf{0.250} & \textbf{0.251} & \textbf{0.252} & \textbf{0.253} & \textbf{0.255} & \textbf{0.256} \\
\midrule
\rowcolor{black!7}\multicolumn{12}{l}{\textbf{Qwen3.6-27B}} \\
Wan2.2-5B & 0.990 & 0.991 & 0.992 & 0.993 & 0.993 & \cellcolor{alphablue!20}0.993 & 0.993 & 0.993 & 0.993 & 0.993 & 0.993 \\
HunyuanVideo-1.5 & 0.953 & 0.957 & 0.959 & 0.961 & 0.963 & \cellcolor{alphablue!20}0.964 & 0.965 & 0.966 & 0.967 & 0.968 & 0.968 \\
LTX-2.3 & 0.988 & 0.990 & 0.991 & 0.991 & 0.992 & \cellcolor{alphablue!20}0.992 & 0.992 & 0.992 & 0.992 & 0.992 & 0.992 \\
Lingbot-Video-Dense & 0.967 & 0.973 & 0.977 & 0.980 & 0.982 & \cellcolor{alphablue!20}0.983 & 0.984 & 0.985 & 0.985 & 0.986 & 0.986 \\
Lingbot-Video-MOE & 0.852 & 0.859 & 0.865 & 0.871 & 0.876 & \cellcolor{alphablue!20}0.881 & 0.886 & 0.891 & 0.895 & 0.897 & 0.900 \\
Minimax H3 & 0.452 & 0.449 & 0.448 & 0.448 & 0.449 & \cellcolor{alphablue!20}0.450 & 0.452 & 0.454 & 0.455 & 0.457 & 0.459 \\
ViduQ3 & 0.932 & 0.937 & 0.940 & 0.943 & 0.945 & \cellcolor{alphablue!20}0.947 & 0.948 & 0.949 & 0.950 & 0.950 & 0.951 \\
HappyHorse1.1 & 0.850 & 0.853 & 0.855 & 0.859 & 0.861 & \cellcolor{alphablue!20}0.863 & 0.865 & 0.867 & 0.868 & 0.870 & 0.870 \\
Kling v3.0 & 0.948 & 0.957 & 0.965 & 0.970 & 0.974 & \cellcolor{alphablue!20}0.977 & 0.978 & 0.980 & 0.981 & 0.981 & 0.982 \\
Seedance2.5 & 0.627 & 0.626 & 0.627 & 0.628 & 0.630 & \cellcolor{alphablue!20}0.632 & 0.634 & 0.635 & 0.636 & 0.638 & 0.640 \\
Wan3.0 & \textbf{0.256} & \textbf{0.252} & \textbf{0.249} & \textbf{0.248} & \textbf{0.248} & \cellcolor{alphablue!20}\textbf{0.248} & \textbf{0.249} & \textbf{0.251} & \textbf{0.252} & \textbf{0.253} & \textbf{0.254} \\
\bottomrule
\end{tabular}%
}
\endgroup
\end{table}

%% file: Sections/Appendix/video_dimensions.tex
\subsection{Video Evaluation by Dimension}
\label{app:video-dimensions}

Table~\ref{tab:video-dimension-results} separates Qwen3.8-27B's Video Scores by dimension. Each score is the proportion of satisfied queries assigned to that dimension, pooled across scenarios; prompts without applicable queries do not contribute. Scene Attributes includes text-carrier presence and integration, whereas Entity Presence covers subjects and objects. The All column pools every query.

\input{Tables/video_dimension_results}

\textbf{The dimension breakdown distinguishes scene fulfillment from dynamic requirements.} Across all 11 models, Scene Attributes, Entity Presence, and Spatial Relationship receive higher scores than Motion Adherence and Temporal Consistency. Wan3.0 leads the first three dimensions, while HappyHorse1.1 achieves the highest Motion Adherence and shares the highest Temporal Consistency with Seedance2.5. The overall score thus combines distinct strengths in scene composition and temporal behavior.

Models with similar overall scores can nevertheless differ in which requirements they fulfill. Kling v3.0 and Seedance2.5 both achieve an overall Video Score of approximately 0.79, yet Kling scores higher on Spatial Relationship, at 0.8118 versus 0.7503, whereas Seedance scores higher on Temporal Consistency, at 0.7049 versus 0.6361. The aggregate score brings these different profiles close together. Reporting individual dimensions makes the distinction explicit and helps identify whether a model's stronger performance lies in arranging scene elements or maintaining the requested behavior over time. This breakdown complements the overall comparison by tying scores to specific generation requirements.

%% file: Tables/video_dimension_results.tex
\begin{table}[!htbp]
\centering
\setlength{\abovecaptionskip}{4pt}
\caption{Scores across five video evaluation dimensions, evaluated by Qwen3.8-27B. Bold values indicate the best result in each column.}
\label{tab:video-dimension-results}
\begingroup
\small
\definecolor{dimensionblue}{HTML}{7390BA}
\setlength{\tabcolsep}{4pt}
\renewcommand{\arraystretch}{1.14}
\resizebox{\linewidth}{!}{%
\begin{tabular}{@{}l|ccccc|c}
\toprule
\textbf{Model} & \multicolumn{6}{c}{\textbf{Score} ($\uparrow$)} \\
\cmidrule(l){2-7}
 & \shortstack{Scene\\Attributes} & \shortstack{Motion\\Adherence} & \shortstack{Spatial\\Relationship} & \shortstack{Entity\\Presence} & \shortstack{Temporal\\Consistency} & All \\
\midrule
\rowcolor{black!7}\multicolumn{7}{l}{\textbf{Open-Sourced Models}} \\
Wan2.2-5B & 0.6308 & 0.1326 & 0.5277 & 0.4813 & 0.0974 & 0.4358 \\
HunyuanVideo-1.5 & 0.7427 & 0.3431 & 0.6519 & 0.7511 & 0.3467 & 0.6013 \\
LTX-2.3 & 0.7279 & 0.3873 & 0.6667 & 0.7076 & 0.3954 & 0.6068 \\
Lingbot-Video-Dense & 0.7068 & 0.2578 & 0.6039 & 0.5487 & 0.3496 & 0.5343 \\
Lingbot-Video-MOE & 0.7485 & 0.2354 & 0.5904 & 0.5787 & 0.3754 & 0.5492 \\
Minimax H3 & 0.8678 & 0.5392 & 0.8032 & 0.8471 & 0.6504 & 0.7562 \\
\midrule
\rowcolor{black!7}\multicolumn{7}{l}{\textbf{Proprietary Models}} \\
ViduQ3 & 0.8720 & 0.6243 & 0.8374 & 0.8586 & 0.6283 & 0.7856 \\
HappyHorse1.1 & 0.8943 & \cellcolor{dimensionblue!25}\textbf{0.7416} & 0.8315 & 0.8666 & \cellcolor{dimensionblue!25}\textbf{0.7049} & 0.8308 \\
Kling v3.0 & 0.8799 & 0.6656 & 0.8118 & 0.8246 & 0.6361 & 0.7930 \\
Seedance2.5 & 0.8752 & 0.6762 & 0.7503 & 0.8396 & \cellcolor{dimensionblue!25}\textbf{0.7049} & 0.7912 \\
Wan3.0 & \cellcolor{dimensionblue!25}\textbf{0.9240} & 0.7285 & \cellcolor{dimensionblue!25}\textbf{0.8672} & \cellcolor{dimensionblue!25}\textbf{0.9070} & 0.7020 & \cellcolor{dimensionblue!25}\textbf{0.8492} \\
\bottomrule
\end{tabular}%
}
\endgroup
\end{table}

%% file: Sections/Appendix/failure_analysis.tex
\subsection{Extended Failure Analysis}
\label{app:failure-analysis}

To examine what makes rendered text difficult to read, we analyze the evaluators' \emph{transcription diagnostics}: one- or two-sentence descriptions of observable difficulties, produced during transcription when an entire target is unreadable or unreadable spans are omitted. These descriptions concern the target as a whole rather than individual word errors.

\begin{figure}[!htbp]
    \setlength{\abovecaptionskip}{4pt}
    \centering
    \subfloat[Diagnostics for unreadable targets.\label{fig:readability-unreadable-words}]{%
        \includegraphics[width=0.495\linewidth]{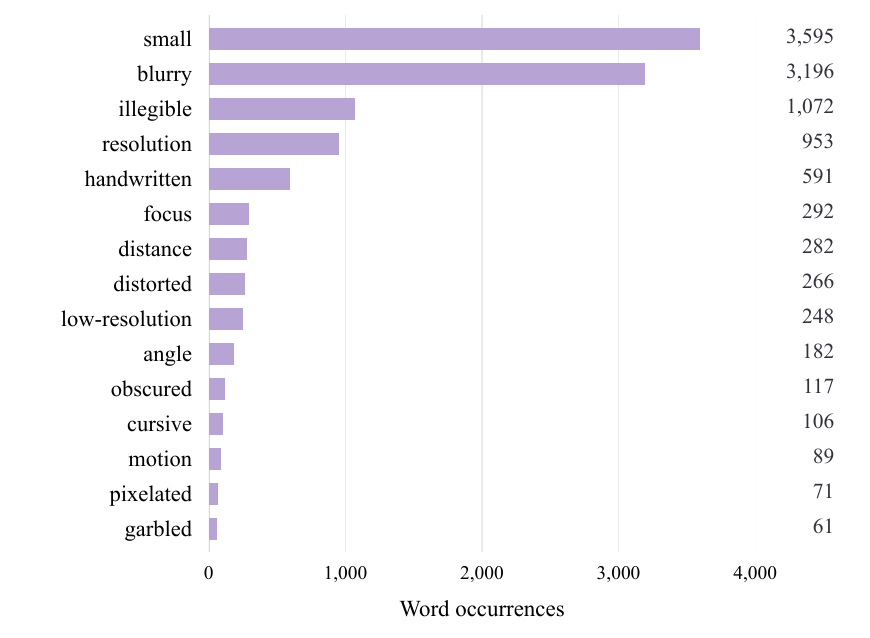}%
    }\hfill
    \subfloat[Diagnostics accompanying transcriptions.\label{fig:readability-transcribed-words}]{%
        \includegraphics[width=0.495\linewidth]{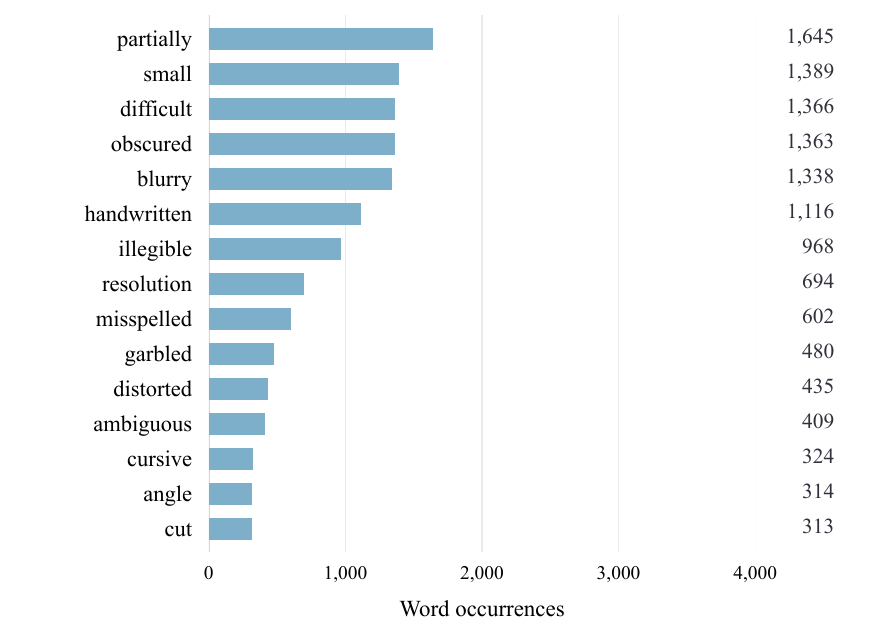}%
    }
    \caption{Word frequencies in transcription diagnostics pooled from Qwen3.8-27B and Qwen3.6-27B. (a) 6,433 descriptions for unreadable targets. (b) 6,549 descriptions accompanying transcriptions. Counts show the 15 most frequent terms within each selected diagnostic vocabulary.}
    \label{fig:readability-analysis}
\end{figure}

Pooling Qwen3.8-27B and Qwen3.6-27B outputs for 13,213 text targets yields 6,433 diagnostics for unreadable targets and 6,549 accompanying transcriptions. Figure~\ref{fig:readability-analysis} shows word occurrences within the selected diagnostic vocabulary. Each evaluator's description is counted separately, and a description may contain multiple terms.

\textbf{Unreadable text is most often associated with insufficient character detail.} In Figure~\ref{fig:readability-unreadable-words}, \emph{small} and \emph{blurry} dominate the diagnostics, with 3,595 and 3,196 occurrences, followed by \emph{illegible} and \emph{resolution}. The vocabulary connects unreadability to the visibility of the rendered characters: a carrier can be recognizable as a sign, document, or interface while its writing lacks enough detail for transcription. Handwriting, viewing distance, and distortion also recur, indicating that the difficulty extends across different carrier appearances and viewing conditions. These observations help explain why the presence of a text-bearing object does not establish that its content can be recovered.

\textbf{Recoverable text combines visibility problems with errors in the rendered content.} Figure~\ref{fig:readability-transcribed-words} is led by \emph{partially}, with 1,645 occurrences, while \emph{obscured} and \emph{blurry} appear 1,363 and 1,338 times. These descriptions capture cases where some text remains accessible despite incomplete visibility or limited detail. At the same time, \emph{misspelled} and \emph{garbled} occur 602 and 480 times. Thus, obtaining a transcription does not imply that the recovered words are well formed or correct. This distinction clarifies why transcription outcomes and WER describe different aspects of the same target: the former records whether text can be recovered, while the latter measures its correspondence to the requested content.

The visual cases illustrate how these difficulties appear in generated scenes. ViduQ3 and Kling v3.0 produce structured interfaces whose prominent text contains pseudo-words. HappyHorse1.1 mixes recognizable fragments with malformed words, while Lingbot-Video-Dense produces abnormal Chinese stroke combinations. The Minimax H3 and LTX-2.3 examples further distinguish distorted handwriting from text that lacks readable detail. Across these cases, the surrounding layout can remain recognizable even when the writing fails at the level of words or character formation. The enlarged crops show how these failures affect the writing on each carrier, linking the visible error to its role within the depicted scene.

\textbf{Content fidelity also requires checking text that appears legible.} The final cases show errors that are directly exposed by the required-text comparison: Wan3.0 replaces $\beta=8/3$ with $\beta=4/3$, Seedance2.5 substitutes a brief update for the requested timing and temperature details, and HunyuanVideo-1.5 renders a prompt instruction fragment as storefront text. These examples connect numerical substitution, information omission, and unintended instruction rendering to the benchmark's explicit textual requirements. Together, the diagnostics and visual cases identify two complementary targets for improvement: making scene text recoverable and preserving the information it is supposed to convey. For applications built around written instructions, records, or numerical specifications, this distinction is consequential: faithfully depicting the carrier is one part of fulfilling the request; the text must also preserve the intended message.

\input{Sections/Appendix/failure_cases}

%% file: Sections/Appendix/failure_cases.tex
\begin{figure}[p]
    \centering
    \captionsetup{font=small}
    \setlength{\abovecaptionskip}{4pt}
    \setlength{\belowcaptionskip}{0pt}
    \includegraphics[width=0.84\linewidth]{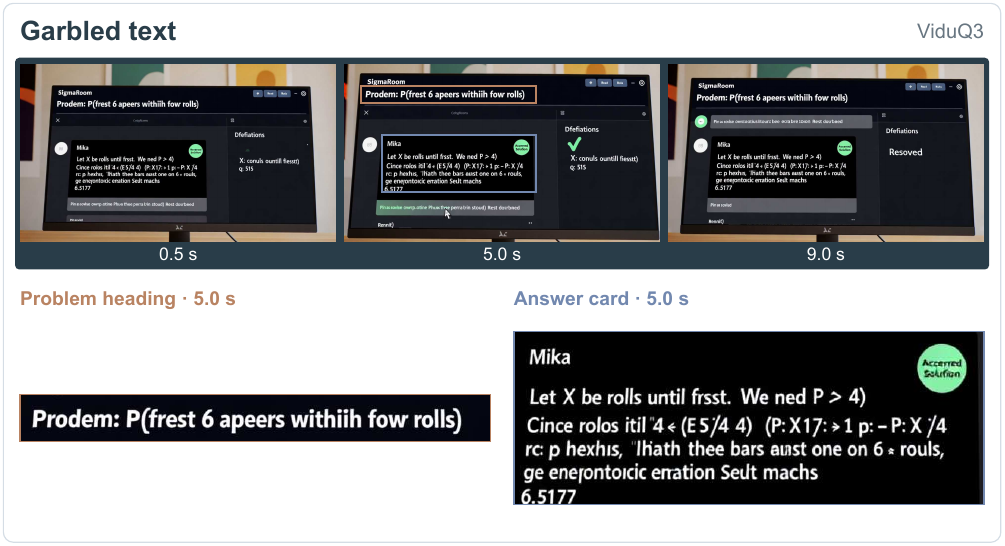}
    \caption{Garbled text in ViduQ3. The problem heading and answer card contain pseudo-words despite their clear appearance.}
    \label{fig:failure-vidu}
    \vspace{8pt}
    \includegraphics[width=0.84\linewidth]{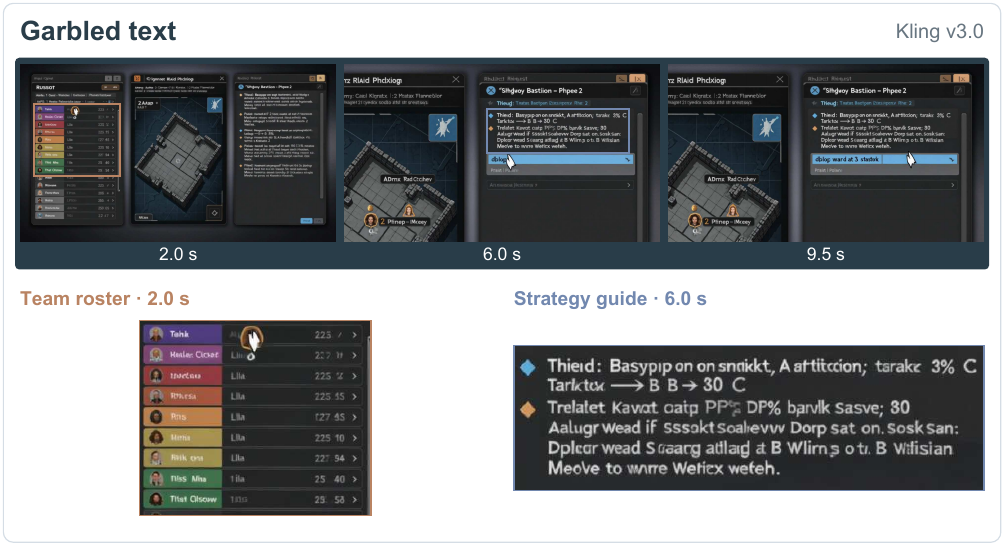}
    \caption{Garbled text in Kling v3.0. The game roster and strategy guide contain extensive pseudo-words within a structured interface.}
    \label{fig:failure-kling}
    \vspace{8pt}
    \includegraphics[width=0.84\linewidth]{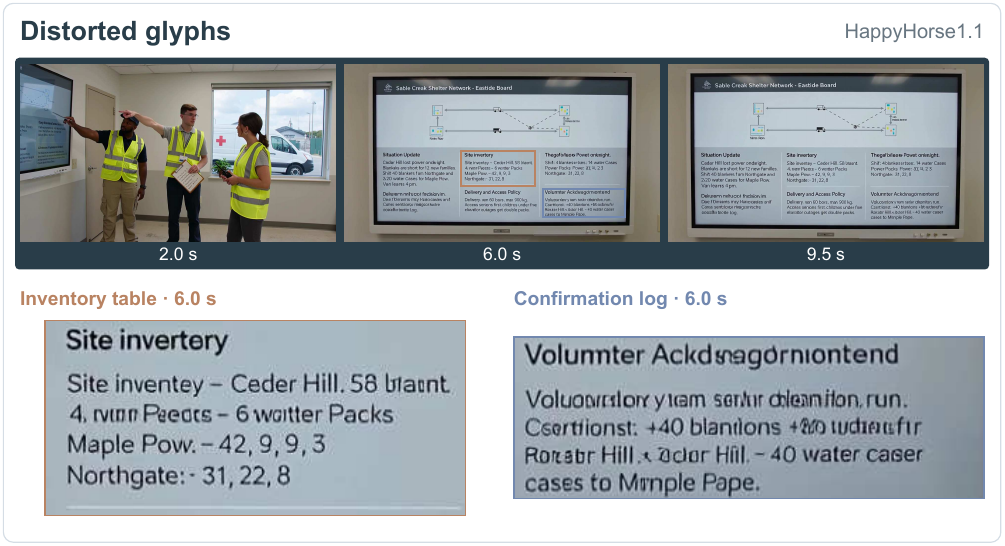}
    \caption{Distorted glyphs in HappyHorse1.1. Inventory and confirmation panels mix readable fragments with malformed words.}
    \label{fig:failure-happyhorse}
\end{figure}
\begin{figure}[p]
    \centering
    \captionsetup{font=small}
    \setlength{\abovecaptionskip}{4pt}
    \setlength{\belowcaptionskip}{0pt}
    \includegraphics[width=0.84\linewidth]{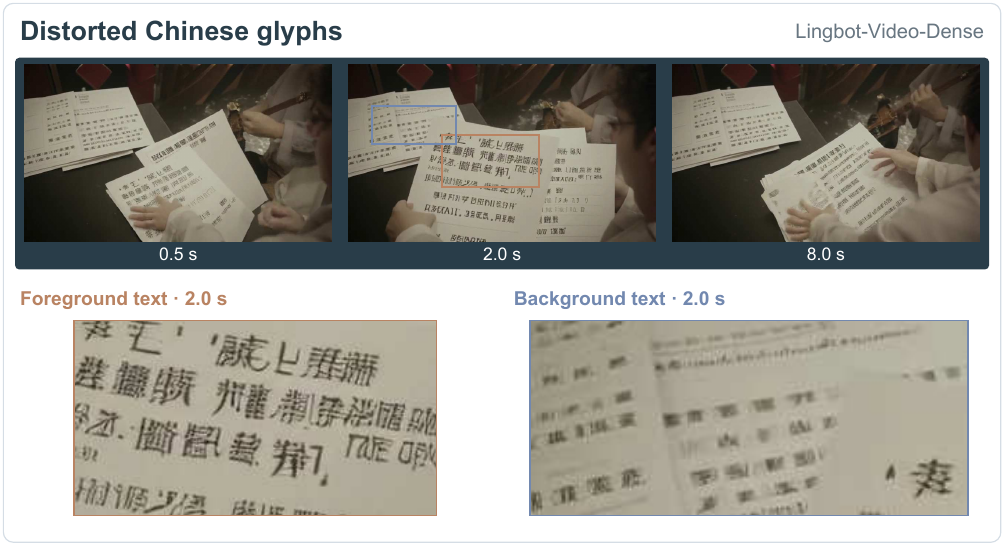}
    \caption{Malformed Chinese characters in Lingbot-Video-Dense. Printed text blocks contain abnormal stroke combinations.}
    \label{fig:failure-lingbot}
    \vspace{8pt}
    \includegraphics[width=0.84\linewidth]{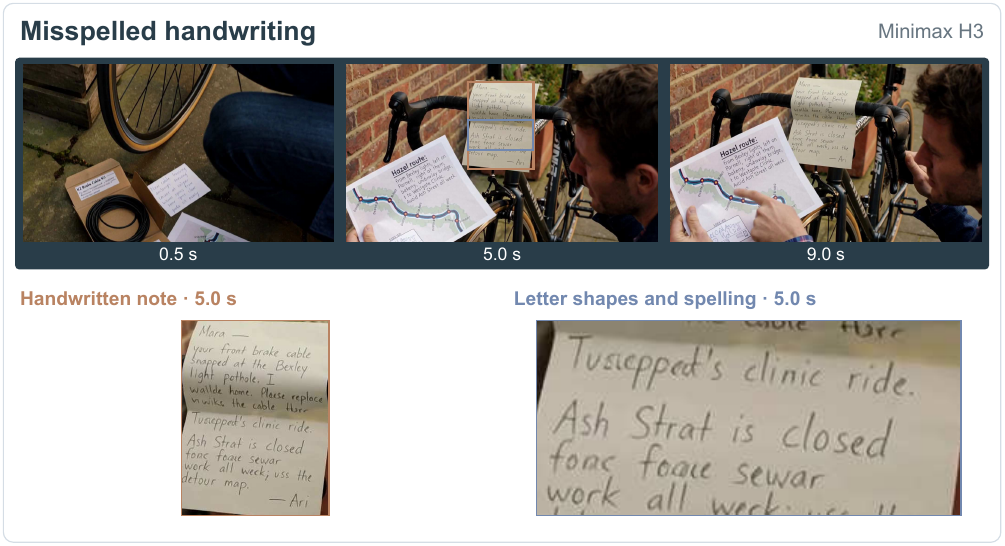}
    \caption{Misspelled handwriting in Minimax H3. The large handwritten note contains distorted word forms.}
    \label{fig:failure-minimax}
    \vspace{8pt}
    \includegraphics[width=0.84\linewidth]{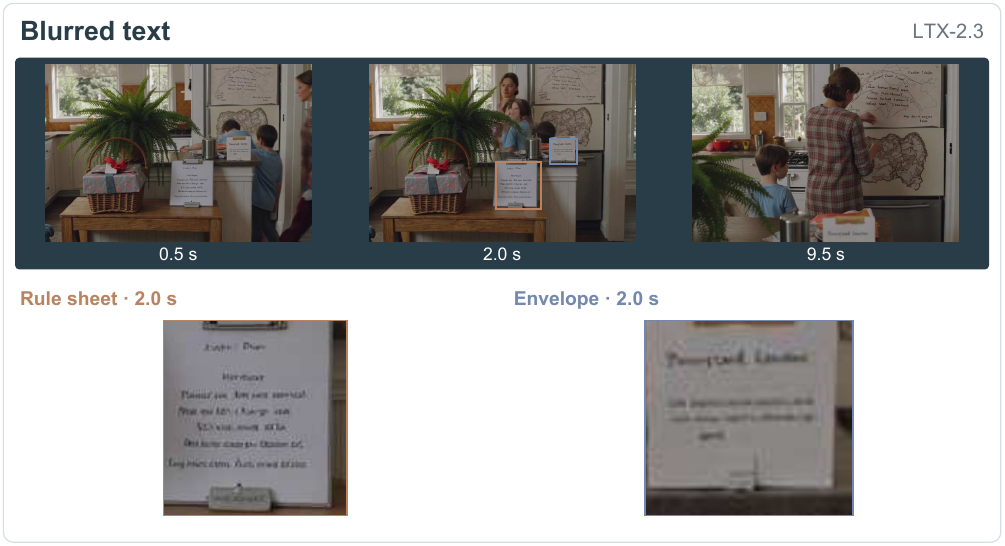}
    \caption{Blurred text in LTX-2.3. The rule sheet and envelope are visible, but their text lacks readable detail.}
    \label{fig:failure-ltx}
\end{figure}
\begin{figure}[p]
    \centering
    \captionsetup{font=small}
    \setlength{\abovecaptionskip}{4pt}
    \setlength{\belowcaptionskip}{0pt}
    \includegraphics[width=0.84\linewidth]{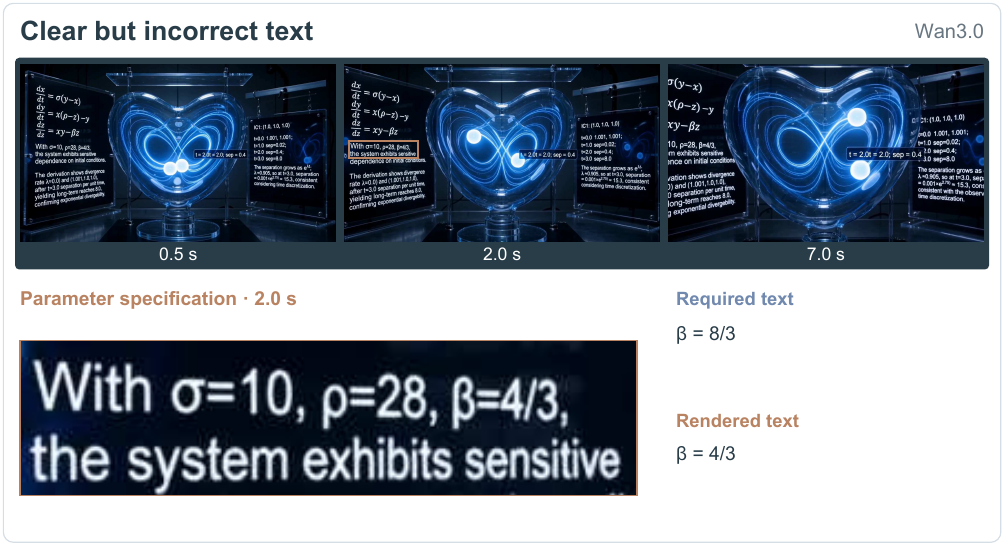}
    \caption{Incorrect parameter in Wan3.0. The required $\beta=8/3$ is rendered as $\beta=4/3$.}
    \label{fig:failure-parameter}
    \vspace{8pt}
    \includegraphics[width=0.84\linewidth]{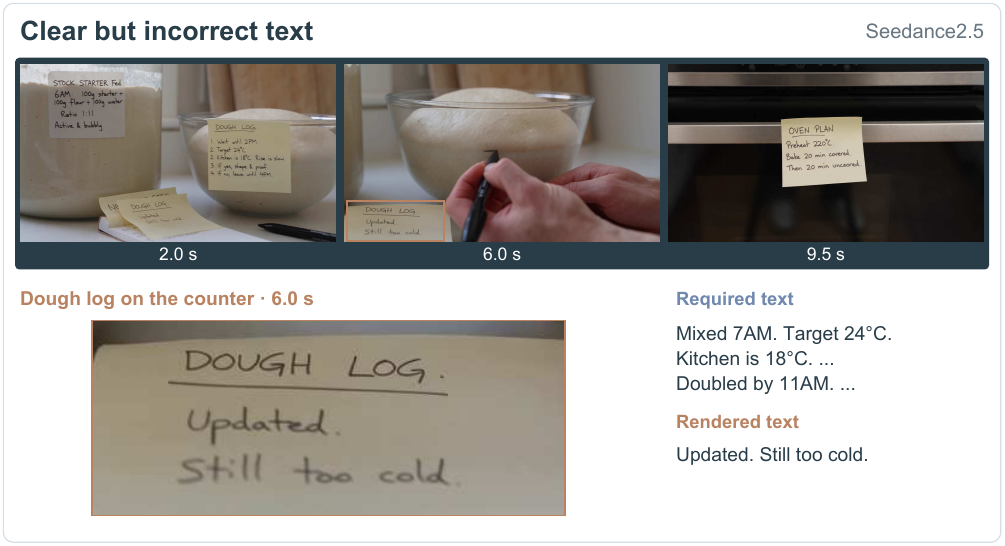}
    \caption{Rewritten information in Seedance2.5. The counter log replaces the required timing and temperature details with a brief update.}
    \label{fig:failure-seedance}
    \vspace{8pt}
    \includegraphics[width=0.84\linewidth]{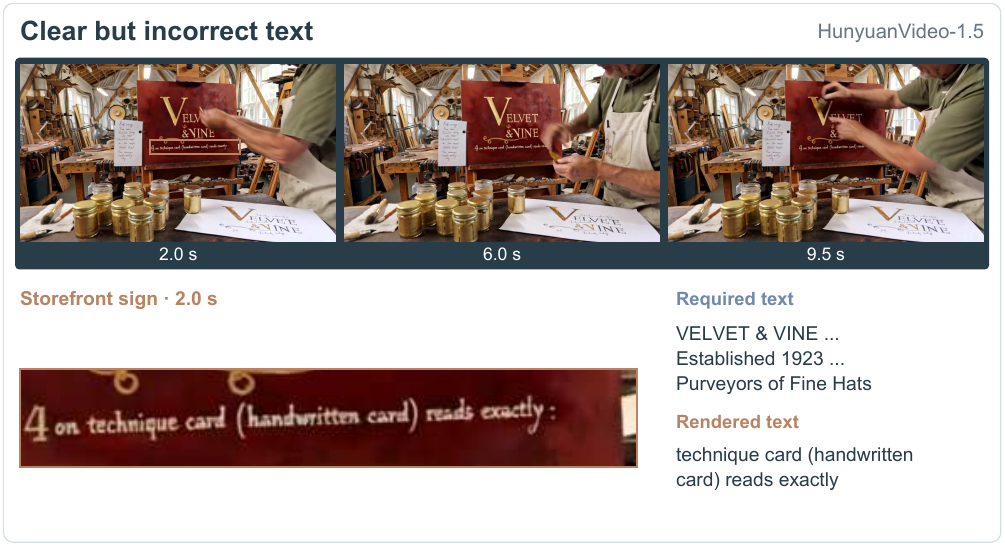}
    \caption{Prompt instructions rendered as visible text in HunyuanVideo-1.5. The storefront sign contains an instruction fragment in place of the required brand copy.}
    \label{fig:failure-hunyuan}
\end{figure}

Figures~\ref{fig:failure-vidu}--\ref{fig:failure-hunyuan} illustrate representative visual text rendering failures.

